\documentclass[journal]{IEEEtran}

\usepackage{times,amsmath,graphicx,subfigure,mathrsfs,mathrsfs,epstopdf,float,threeparttable,hyperref}
\usepackage{xcolor,array}
\usepackage{soul}
\usepackage[utf8]{inputenc}
\usepackage[small]{caption}

\ifCLASSINFOpdf
\else
\fi

\begin{document}

\title{Lightweight Pyramid Networks for Image Deraining}

\author{Xueyang Fu, Borong Liang, Yue Huang, Xinghao Ding* and John Paisley
\thanks{This work was supported in part by the National Natural Science Foundation of China grants 61571382, 81671766, 61571005, 81671674, U1605252, 61671309 and 81301278, Guangdong Natural Science Foundation grant 2015A030313007, Fundamental Research Funds for the Central Universities grants 20720160075 and 20720150169, the Natural Science Foundation of Fujian Province of China grant 2017J01126, and the CCF-Tencent research fund.  \textit{ (*Corresponding author: Xinghao Ding, dxh@xmu.edu.cn.)}}
\thanks{X. Fu, B. Liang, Y. Huang and X. Ding  are with Fujian Key Laboratory of Sensing and Computing for Smart City, School of Information Science and Engineering, Xiamen University, Xiamen 361005, China.}
\thanks{J. Paisley is with the Department of Electrical Engineering \& Data Science Institute, Columbia University, New York, NY 10027 USA.}
}

\maketitle

\begin{abstract}
Existing deep convolutional neural networks have found major success in image deraining, but at the expense of an enormous number of parameters. This limits their potential application, for example in mobile devices. In this paper, we propose a lightweight pyramid of networks (LPNet) for single image deraining. Instead of designing a complex network structures, we use domain-specific knowledge to simplify the learning process. Specifically, we find that by introducing the mature Gaussian-Laplacian image pyramid decomposition technology to the neural network, the learning problem at each pyramid level is greatly simplified and can be handled by a relatively shallow network with few parameters. We adopt recursive and residual network structures to build the proposed LPNet, which has less than 8K parameters while still achieving state-of-the-art performance on rain removal. We also discuss the potential value of LPNet for other low- and high-level vision tasks.
\end{abstract}


\section{Introduction}
\begin{figure}[!h]
\begin{center}
\subfigure[Rainy image]{\includegraphics[width = 1.7in]{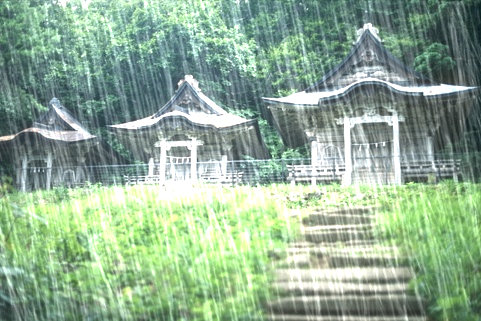}}
\subfigure[Our result]{\includegraphics[width = 1.7in]{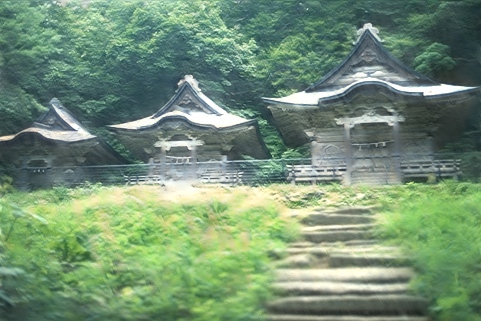}}
\caption{An deraining example of our LPNet for single image deraining. The whole network only contains \textbf{7,548} parameters.} \label{fig.exmple1}
\end{center}
\end{figure}

As a common weather condition, rain impacts not only human visual perception but also computer vision systems, such as self driving vehicles and surveillance systems. Due to the effects of light refraction and scattering, objects in an image are easily blurred and blocked by individual rain streaks. When facing heavy rainy conditions, this problem becomes more severe due to the increased density of rain streaks. Since most existing computer vision algorithms are designed based on the assumption of clear inputs, their performance is easily degraded by rainy weather. Thus, designing effective and efficient algorithms for rain streak removal is a significant problem with many downstream uses. Figure \ref{fig.exmple1} shows an example of our lightweight pyramid network.

\subsection{Related works}

\begin{figure*}[!htp]
\begin{center}
\includegraphics[width = .98\textwidth]{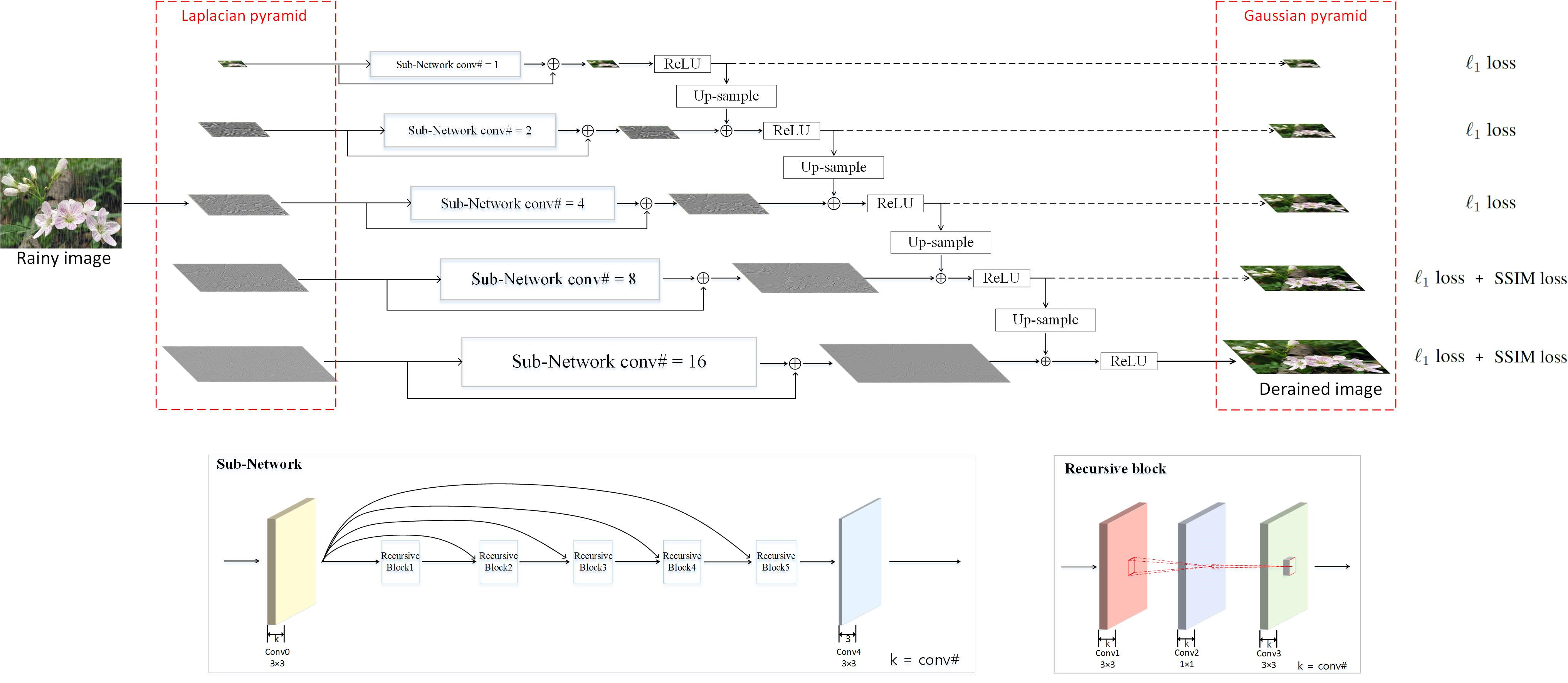}
\caption{The proposed structure of our deep lightweight pyramid of networks (LPNet) based on Gaussian-Laplacian image pyramids. The bottom level of the reconstructed Gaussian pyramid is the final de-rained image.} \label{fig.flowchart}
\end{center}
\end{figure*}

Depending on the input data, rain removal algorithms can be categorized into video and single-image based methods.

\subsubsection{Video based methods}
We first briefly review the rain removal methods in a video, which was the major focus in the early stages of this problem. These methods use both spatial and temporal information from video.
The first study on video deraining removed rain from a static background using average intensities from the neighboring frames \cite{garg2004detection}. Other methods focus on deraining in the Fourier domain \cite{barnum2010analysis}, using Gaussian mixture models \cite{bossu2011rain}, low rank approximations \cite{chen2013generalized} and via matrix completions \cite{kim2015video}. In \cite{ren2017video}, the authors divide rain streaks into sparse ones and dense ones, then a matrix decomposition based algorithm is proposed for deraining. More recently, \cite{wei2017should} proposed a patch-based mixture of Gaussians for rain removal in video. Though these methods work well, they require temporal content of video. In this paper we instead focus on the single image deraining problem.

\subsubsection{Single-image methods}
Since information is drastically reduced in individual images, single image deraining is a much more difficult problem. Methods for addressing this problem have employed kernels \cite{kim2013single}, low rank approximations \cite{chen2013generalized,chang2017transformed} and dictionary learning \cite{Kang2012automatic,Huang2014Self,Luo2015Removing,Wang2017Hierarchical}. In \cite{kim2013single}, rain streaks are detected and removed by using kernel regression and a non-local mean filtering. In \cite{Kang2012automatic}, the authors decompose a rainy image into its low- and high- frequency components. The high-frequency part is processed to extract and remove rain streaks by using sparse-coding based dictionary learning. In \cite{Huang2014Self}, a self learning method is proposed to automatically distinguish rain streaks from the high-frequency part. A discriminative sparse coding method is proposed in \cite{Luo2015Removing}. By forcing the coefficient vector of rain layer to be sparse, the objective function is solved to separate background and rain streaks. Other methods have used mixture models \cite{Li2016Rain} and local gradients \cite{zhu2017joint} to model and then remove rain streaks. In \cite{Li2016Rain}, by utilizing Gaussian Mixture Models (GMMs), the authors explore patch-based priors for both the clean and rain layers. The GMM prior for background layers is learned from natural images, while that for rain streaks layers is learned from rainy images. In \cite{zhu2017joint}, three new priors are defined by exploring local image gradients. The priors are used to modeling the objective function which is solved by using alternating direction method of multipliers (ADMM).

Deep learning has also been introduced for this problem. Convolutional neural networks (CNN) have proven useful for a variety of high-level vision tasks \cite{krizhevsky2012imagenet,Cecotti2014Single,Chen2016DISC,Gong2016Change,he2016deep} as well as various image processing problems \cite{hou2015blind,dong2016image,tai2017memnet,hu2017memristive,dian2018deep}. In \cite{eigen2013restoring}, a related work based on deep learning was introduced to remove static raindrops and dirt spots from pictures taken through windows. Our previous CNN-based method for removing dynamic rain streaks was introduced by \cite{fu2017clearing}. Here the authors build a relative shallow network with 3 layers to extract features of rain streaks from the high frequency content of a rainy image. Based on the introduction of an effective strategy for training very deep networks \cite{he2016deep}, two deeper networks were proposed based on image residuals \cite{fu2017removing} and multi-scale information \cite{Yang2017Deep}. In \cite{zhang2017image}, the authors utilize the generative adversarial framework to further enhance the textures and improve the visual quality of de-rained results. Recently, in \cite{zhang2018density}, a density aware multi-stream densely connected CNN is proposed for joint rain density estimation and de-raining. This method can automatically generate rain density label, which is further utilized to guide rain streaks removal.

\subsection{Our contributions}
Though very deep networks achieve excellent performance on single image deraining, a main drawback that potentially limits their application in mobile devices, automatic driving, and other computer vision tasks is their huge number of parameters. As a networks become deeper, more storage space is required \cite{han2016deep}. To address this issue, we propose a lightweight pyramid network (LPNet), which contains fewer than 8K parameters, with the single image rain removal problem in mind. Instead of designing a complex network structure, we use problem-specific knowledge to simplify the learning process. Specifically, we first adopt Laplacian pyramids to decompose a degraded/rainy image into different levels. Then we use recursive and residual networks to build a sub-network for each level to reconstruct Gaussian pyramids of derained images. A specific loss function is selected for training each sub-network according to its own physical characteristics and the whole training is performed in a multi-task supervision. The final recovered image is the bottom level of the reconstructed Gaussian pyramid.

The main feature of our LPNet approach is to use the mature Gaussian-Laplacian image pyramid technique \cite{Burt1983laplacian} to transform one hard problem into several easier sub-problems. In other words, since the Laplacian pyramid contains different levels that can differentiate large scale edges from small scale details, one can design simple and lightweight sub-network to handle each level in a divide-and-conquer way. The contributions of our paper are summarized as follows:
\begin{enumerate}
\item We show how by combining the classical Gaussian-Laplacian pyramid technique with CNN, a simple network structure with few parameters and relative shallow depth is sufficient for excellent performance. To our knowledge the resulting network is far more lightweight (in terms of parameters) among deep networks with comparable performance.
\item Through multi-scale techniques and recursive and residual deep learning, our proposed network achieves state-of-the-art performances on single image deraining. Although LPNet is trained on synthetic data by necessity, it still generalizes well to real-world images.
\item We discuss how LPNet can be applied to other fundamental low- and high-level vision tasks in image processing. We also show how LPNet can improve downstream applications such as object recognition.
\end{enumerate}

\section{Lightweight pyramid network for deraining}
In Figure \ref{fig.flowchart}, we show our proposed LPNet for single image deraining. To summarize at a high level, we first decompose a rainy image into a Laplacian pyramid and build a sub-network for each pyramid level. Then each sub-network is trained with its own loss function according to the specific physical characteristics of the data at that level. The network outputs a Gaussian pyramid of the derained image. The final derained result is the bottom level of the Gaussian pyramid .

\subsection{Motivation}
Since rain streaks are blended with object edges and the background scene, it is hard to directly learn the deraining function in the image domain \cite{fu2017clearing}. To simplify the problem, it is natural to train a network on the high-frequency information in images, which primarily contain rain streaks and edges without background interference. Based on this motivation, the authors in \cite{fu2017clearing,fu2017removing} use the guided filter \cite{he2013Guided} to obtain the high-frequency component of an image as the input to a deep network, which is then derained and fused back with the low-resolution information of the same image. However, these two methods fail when very thick rain streaks cannot be extracted by the guided filter. Inspired by this decomposition idea, we instead build a lightweight pyramid of networks to instead simplify the learning processing and reduce the number of necessary parameters as a result.

\subsection{Stage 1: The Laplacian pyramid}
We first decompose a rainy image $\bf{X}$ into its Laplacian pyramid, which is a set of images $L$ with $N$ levels:
\begin{align}\label{eq.lap}
{L_{n}}({\bf{X}})  = {G_n}({\bf{X}})- \text{upsample}({G_{n+1}}({\bf{X}})),
\end{align}
where ${G_n}$ is the Gaussian pyramid, $n = 1,...,N-1$. The function ${G_{n}}({\bf{X}})$ is computed by downsampling ${G_{n-1}}(\bf{X})$ using a Gaussian kernel, with ${G_1}(\bf{X}) = \bf{X}$ and ${L_N}({\bf{X}}) = {G_N}(\bf{X})$.

\begin{figure}
\begin{center}
\subfigure[\scriptsize{Rainy image $\bf{X}$}]{\includegraphics[width = 0.82in,height = 0.53in]{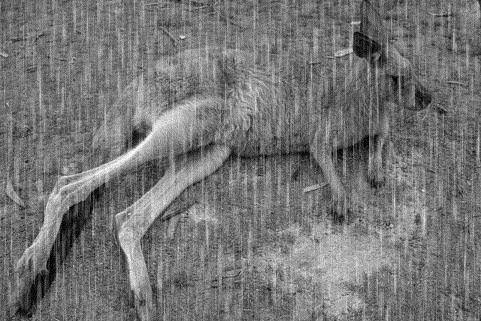}}
\subfigure[\scriptsize{$L_1(\bf{X})$}]{\includegraphics[width = 0.82in,height = 0.53in]{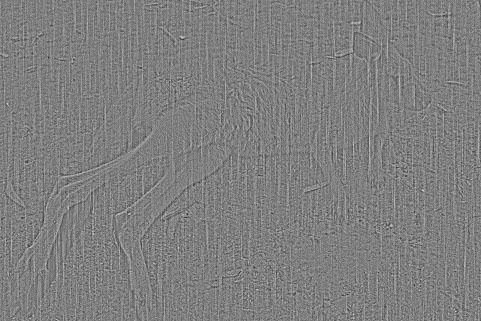}}
\subfigure[\scriptsize{$L_3(\bf{X})$}]{\includegraphics[width = 0.82in,height = 0.53in]{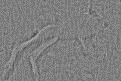}}
\subfigure[\scriptsize{$L_5(\bf{X})$}]{\includegraphics[width = 0.82in,height = 0.53in]{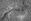}} \\ \vspace{-0.1in}
\subfigure[\scriptsize{Clean image $\bf{Y}$}]{\includegraphics[width = 0.82in,height = 0.53in]{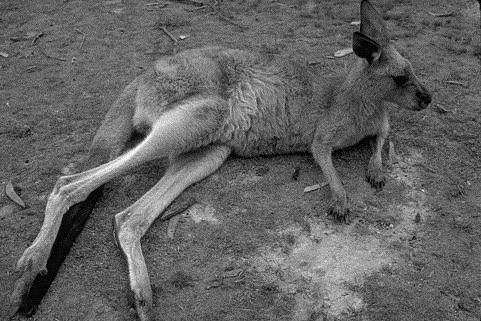}}
\subfigure[\scriptsize{$L_1(\bf{Y})$}]{\includegraphics[width = 0.82in,height = 0.53in]{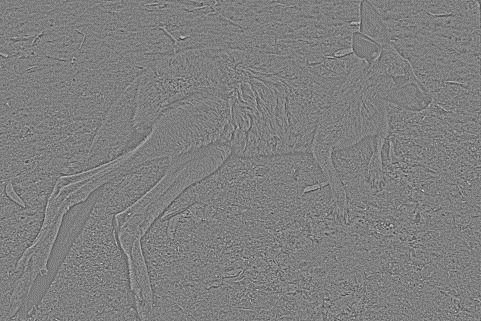}}
\subfigure[\scriptsize{$L_3(\bf{Y})$}]{\includegraphics[width = 0.82in,height = 0.53in]{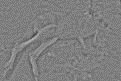}}
\subfigure[\scriptsize{$L_5(\bf{Y})$}]{\includegraphics[width = 0.82in,height = 0.53in]{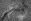}} \\  \vspace{-0.1in}
\subfigure[\scriptsize{(a) - (e)}]{\includegraphics[width = 0.82in,height = 0.53in]{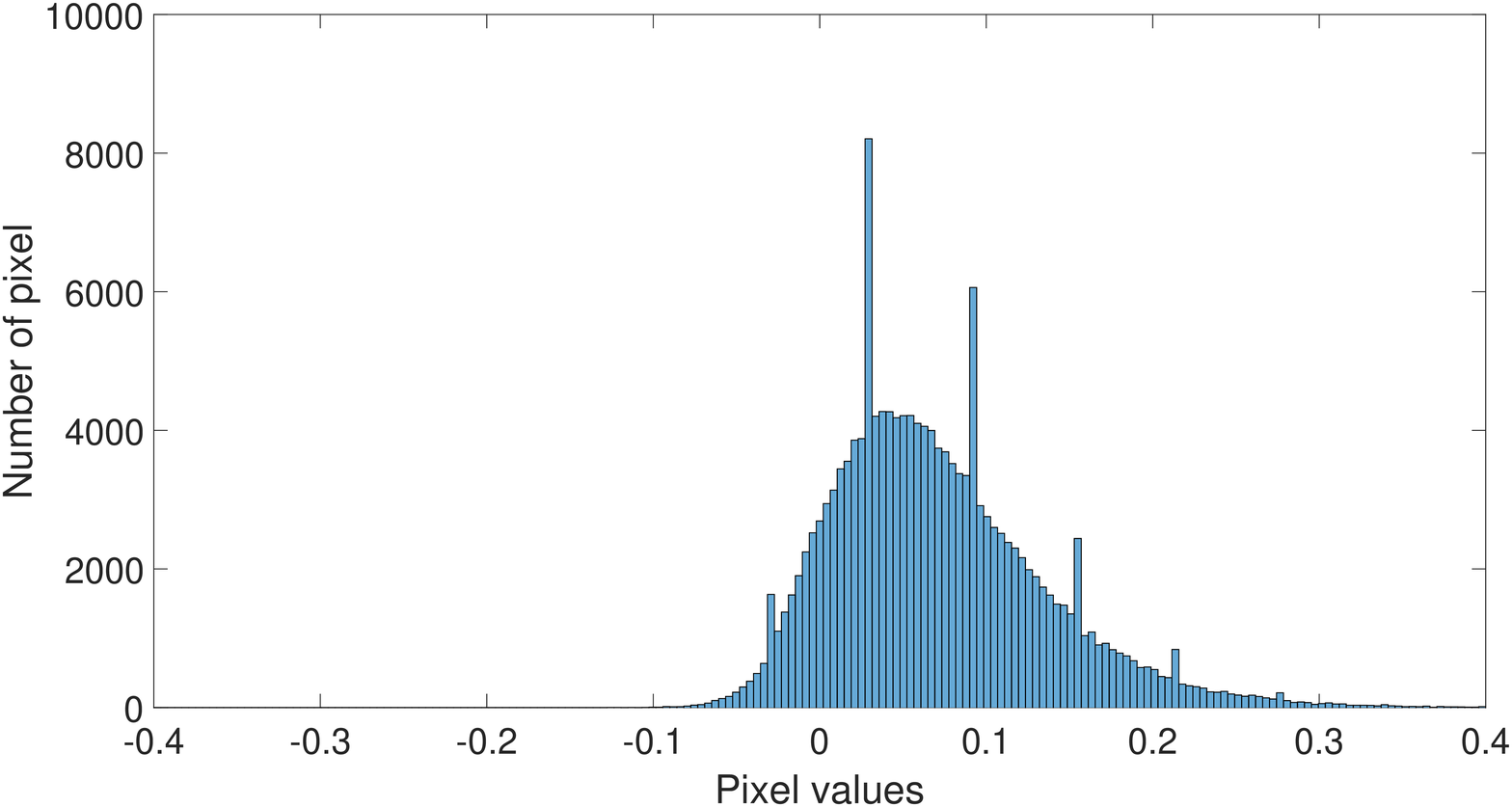}}
\subfigure[\scriptsize{(b) - (f)}]{\includegraphics[width = 0.82in,height = 0.53in]{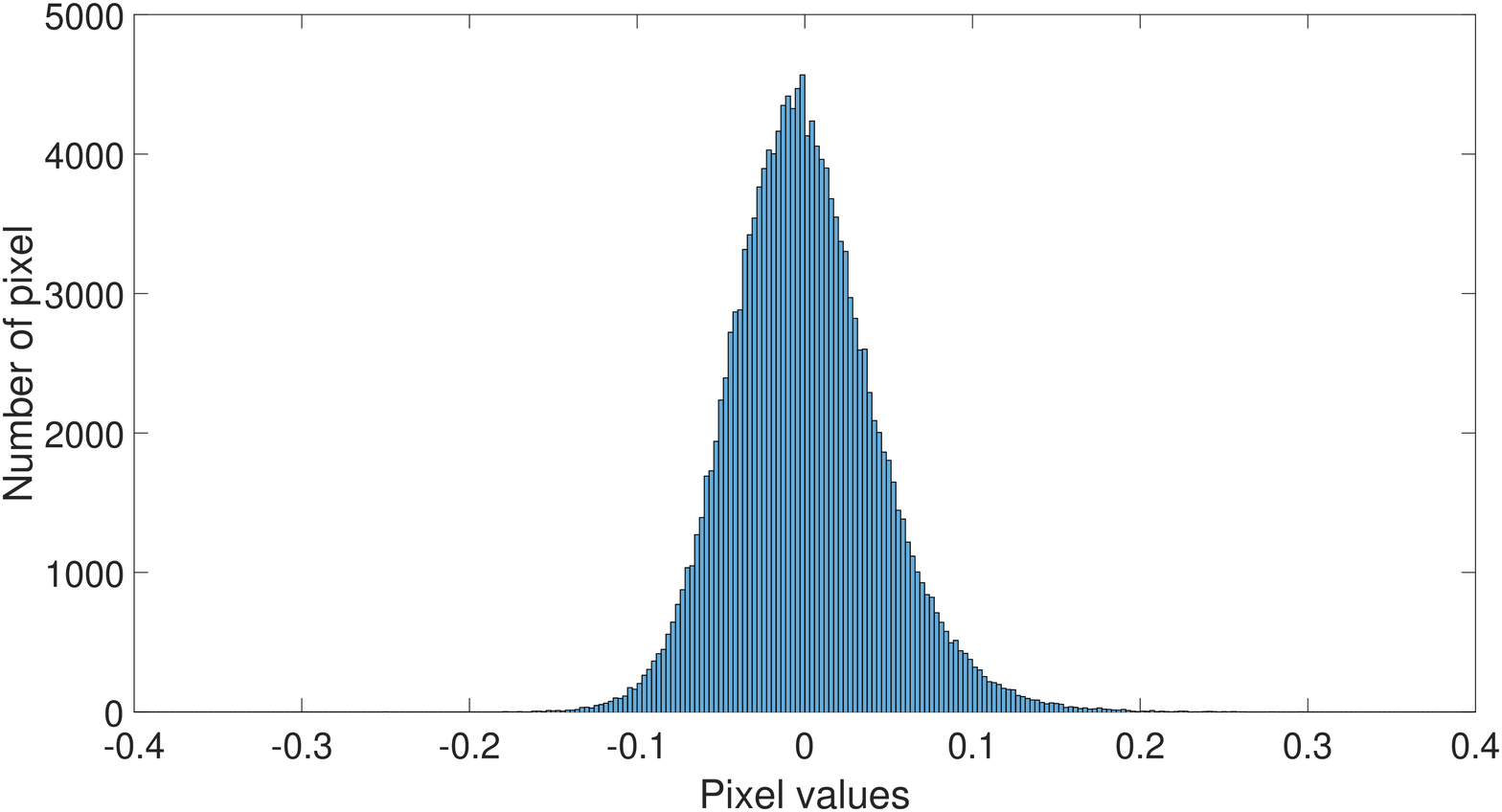}}
\subfigure[\scriptsize{(c) - (g)}]{\includegraphics[width = 0.82in,height = 0.53in]{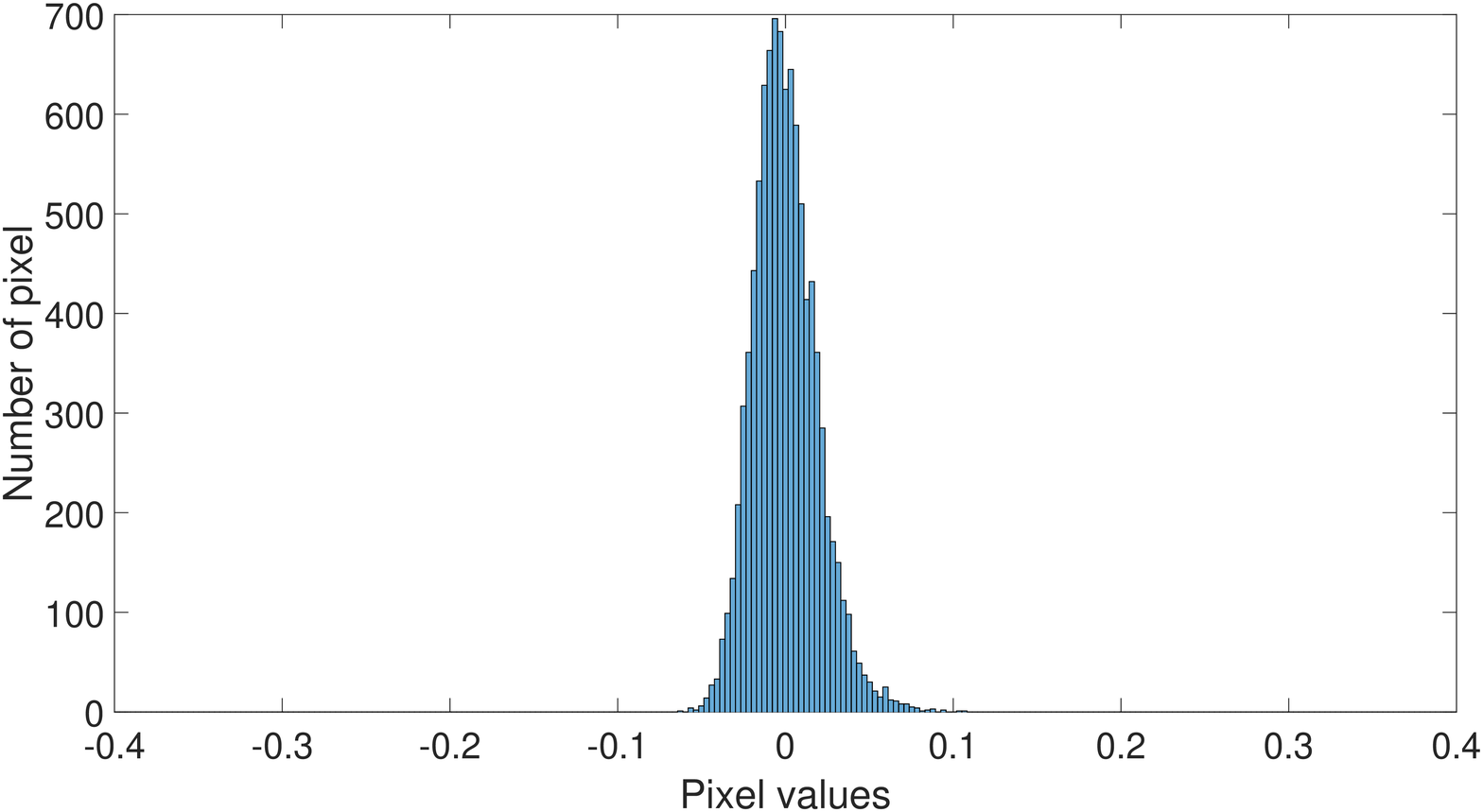}}
\subfigure[\scriptsize{(d) - (h)}]{\includegraphics[width = 0.82in,height = 0.53in]{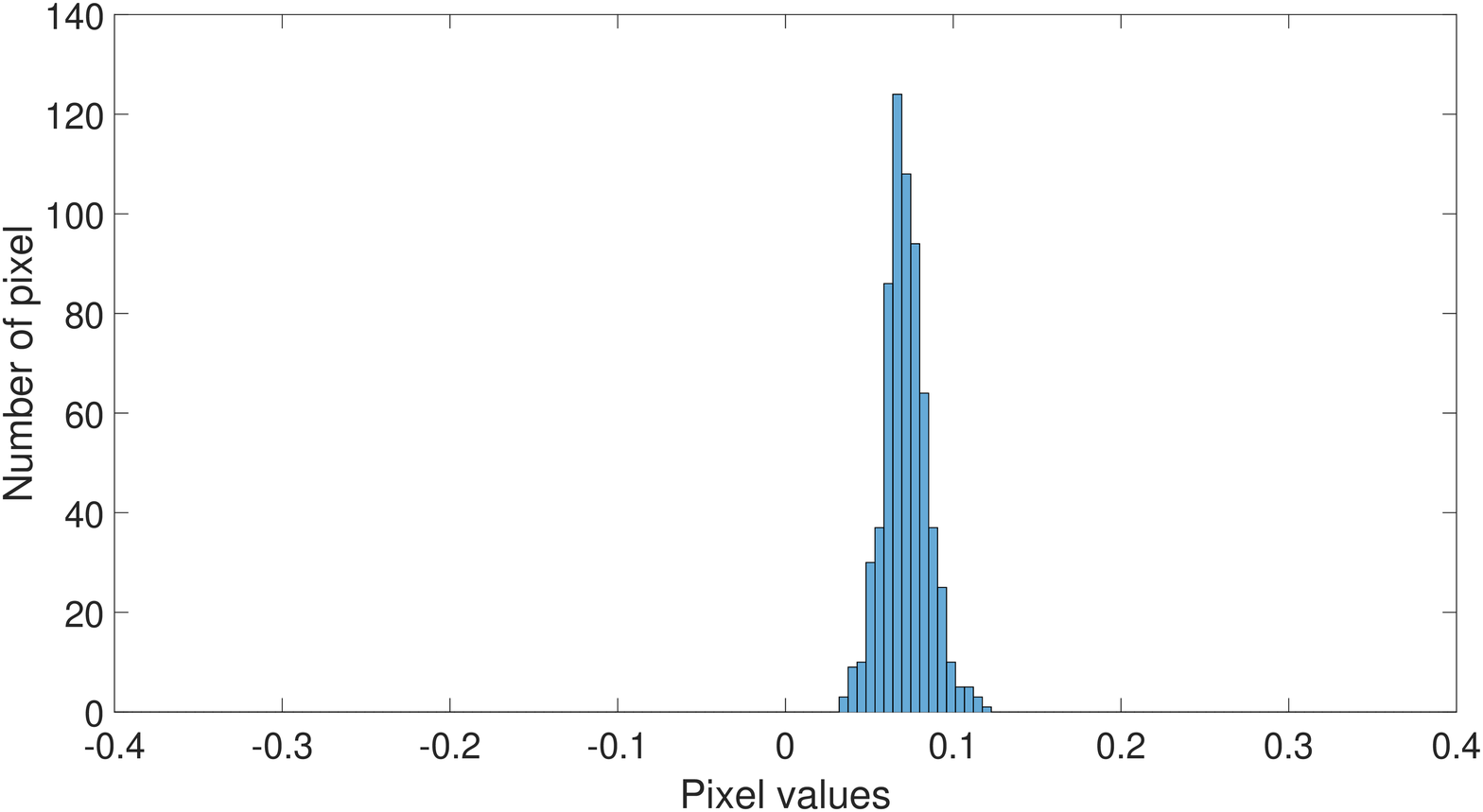}}
\caption{An example of Laplacian pyramid. We show three levels here. The 3rd and 5th level are increased in size for better visualization. The bottom row shows the histogram of the residual to demonstrate the increased sparsity over the image domain.} \label{fig.hist}
\end{center}
\end{figure}

The reasons we choose the classical Laplacian pyramid to decompose the rainy image are fourfold: 1) The background scene can be fully extracted at the top level of ${L_n}$ while the other levels contain rain streaks and details at different spatial scales. Thus, the rain interference is removed and each sub-network only needs to deal with high-frequency components at a single scale. 2) This decomposition strategy will allow the network to take advantage of the sparsity at each level, which motivates many other deraining methods \cite{kim2013single,Huang2014Self,fu2017clearing}, to simplify the learning problem. However, unlike previous deraining methods that use a single-scale decomposition, LPNet performs a multi-scale decomposition using Laplacian pyramids. 3) As shown in Figure \ref{fig.hist}, compared with the image domain, deep learning at each pyramid level is more like an identity mapping (e.g., the top row is more similar to the middle row, as evident in the bottom row) which is known to be the situation where residual learning (ResNet) excels \cite{he2016deep}. 4) The Laplacian pyramid is a mature algorithm with low computation cost. Most calculations are based on convolutions (Gaussian filtering) which can be easily embedded into existing systems with GPU acceleration.

\begin{figure*}[!htp]
\begin{center}
\subfigure[Image domain]{\includegraphics[width = 2.1in]{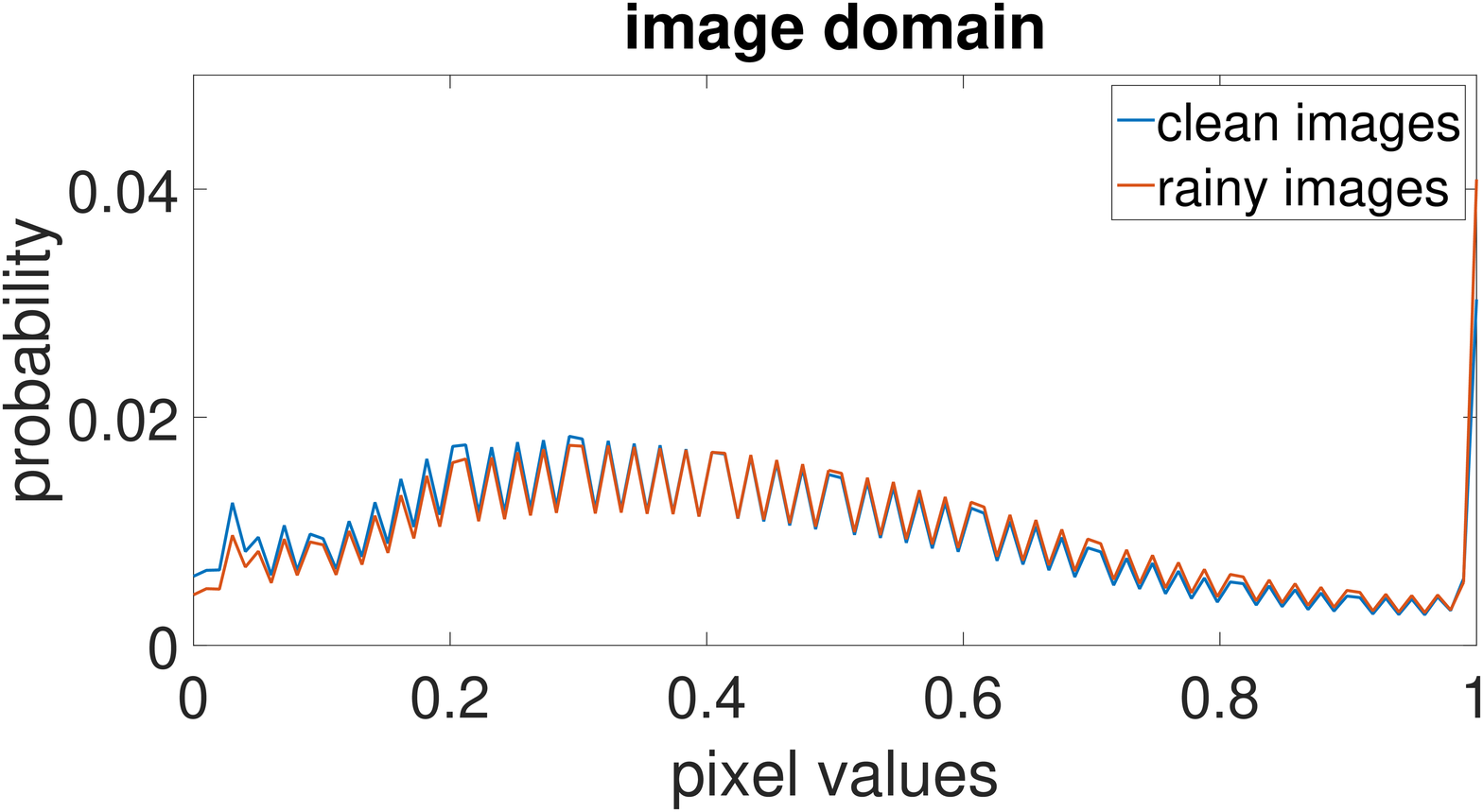}}
\subfigure[5th level]{\includegraphics[width = 2.1in]{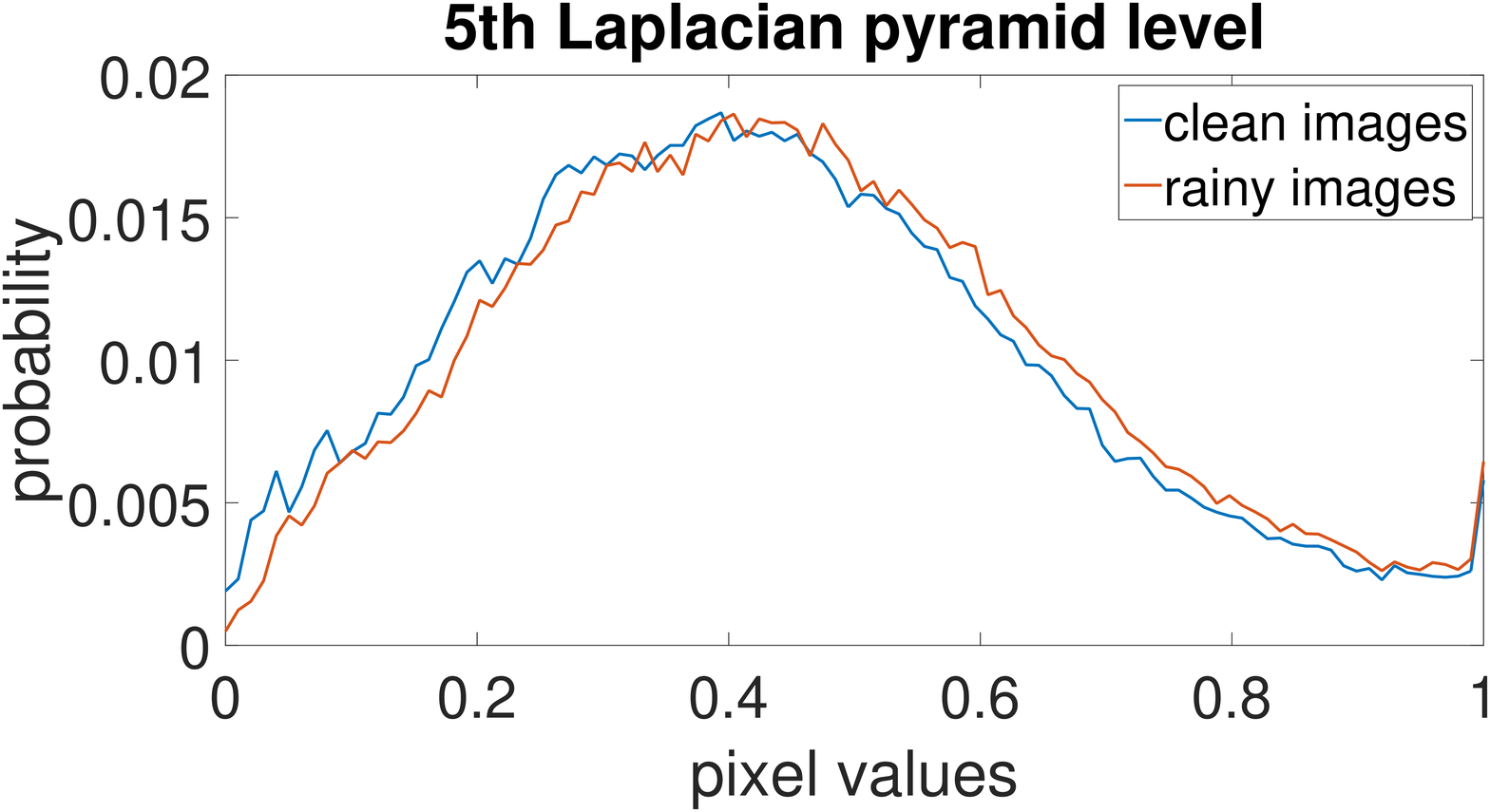}}
\subfigure[4th level]{\includegraphics[width = 2.1in]{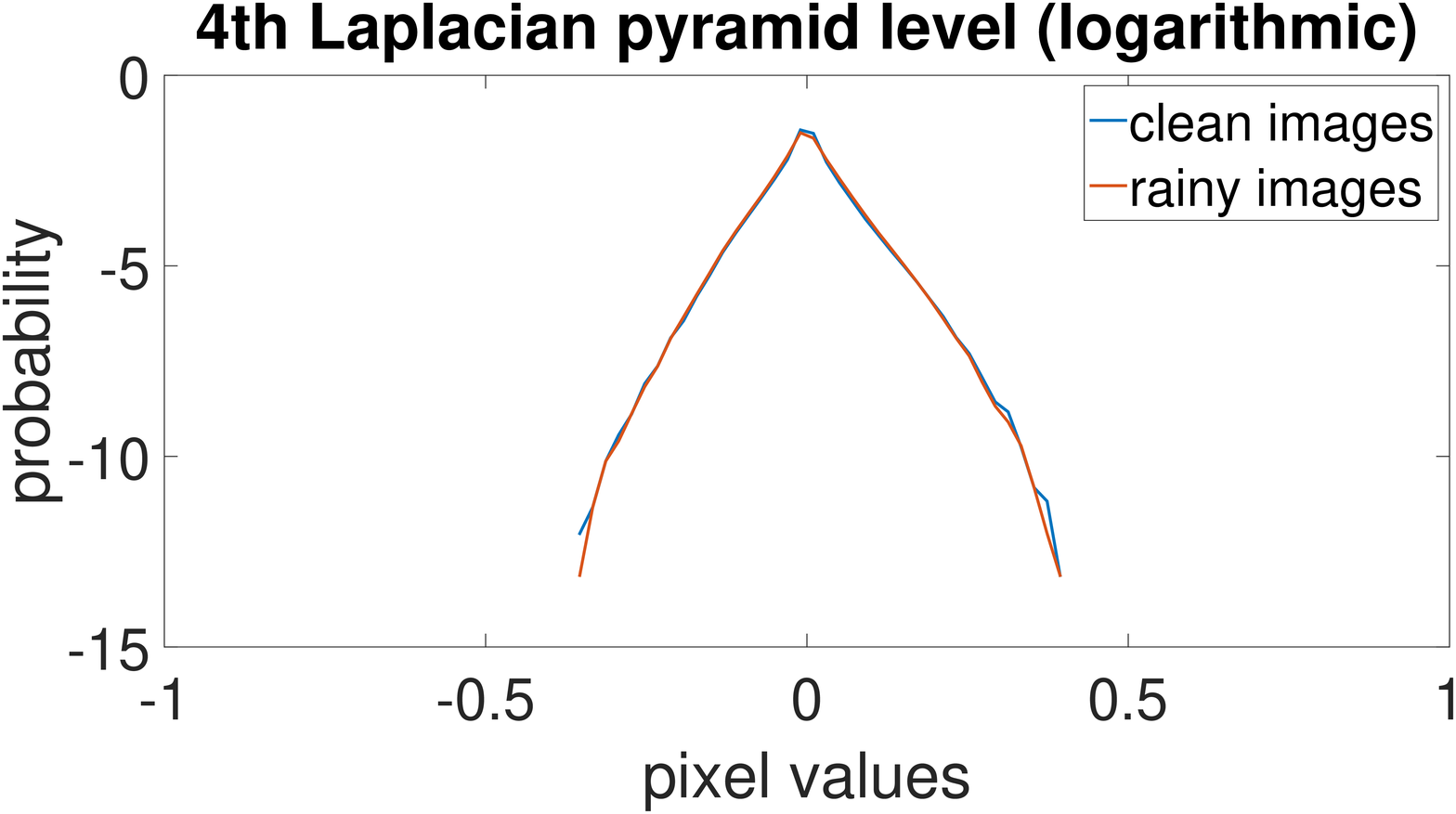}}\\
\subfigure[3rd level]{\includegraphics[width = 2.1in]{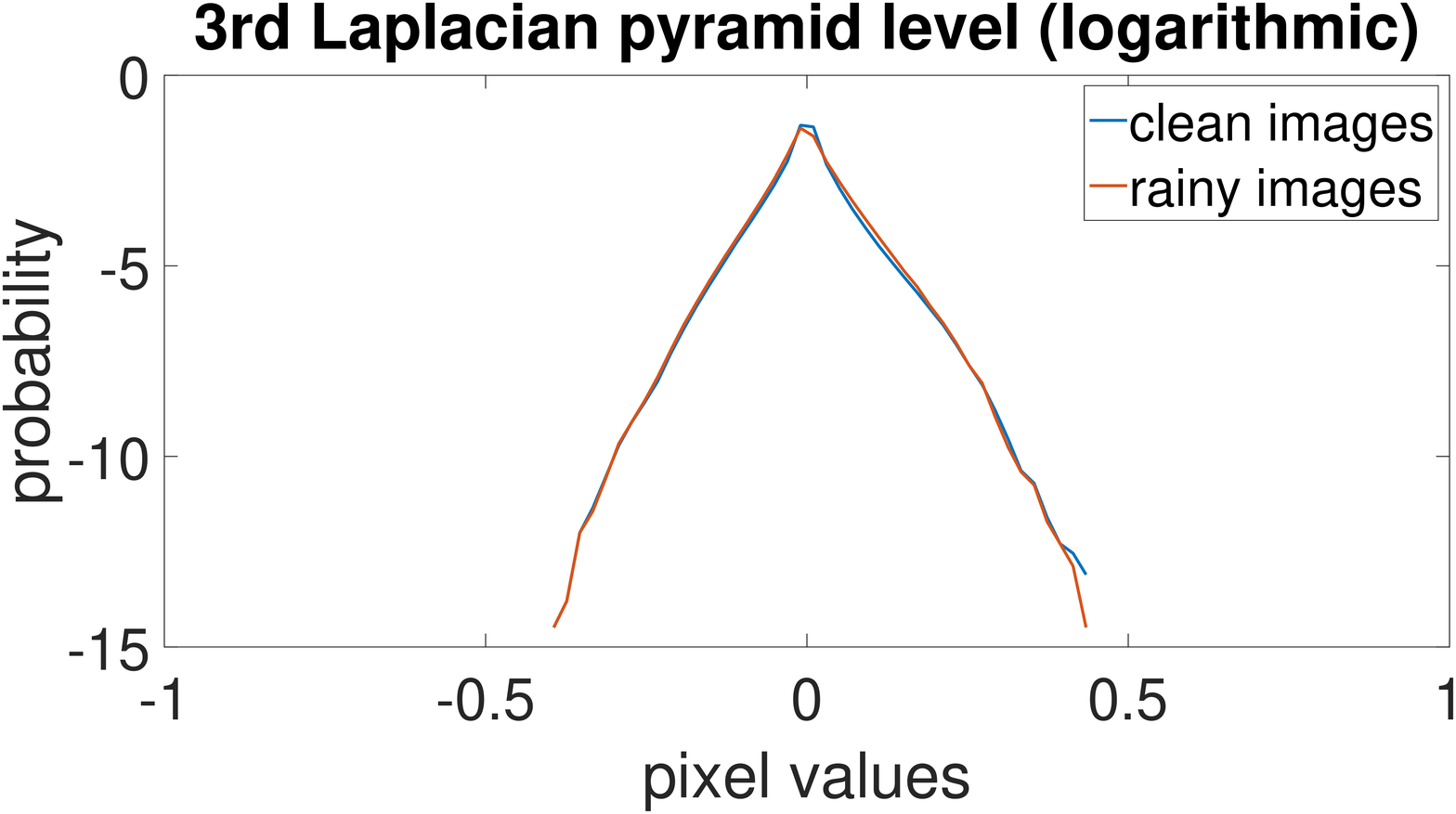}}
\subfigure[2nd level]{\includegraphics[width = 2.1in]{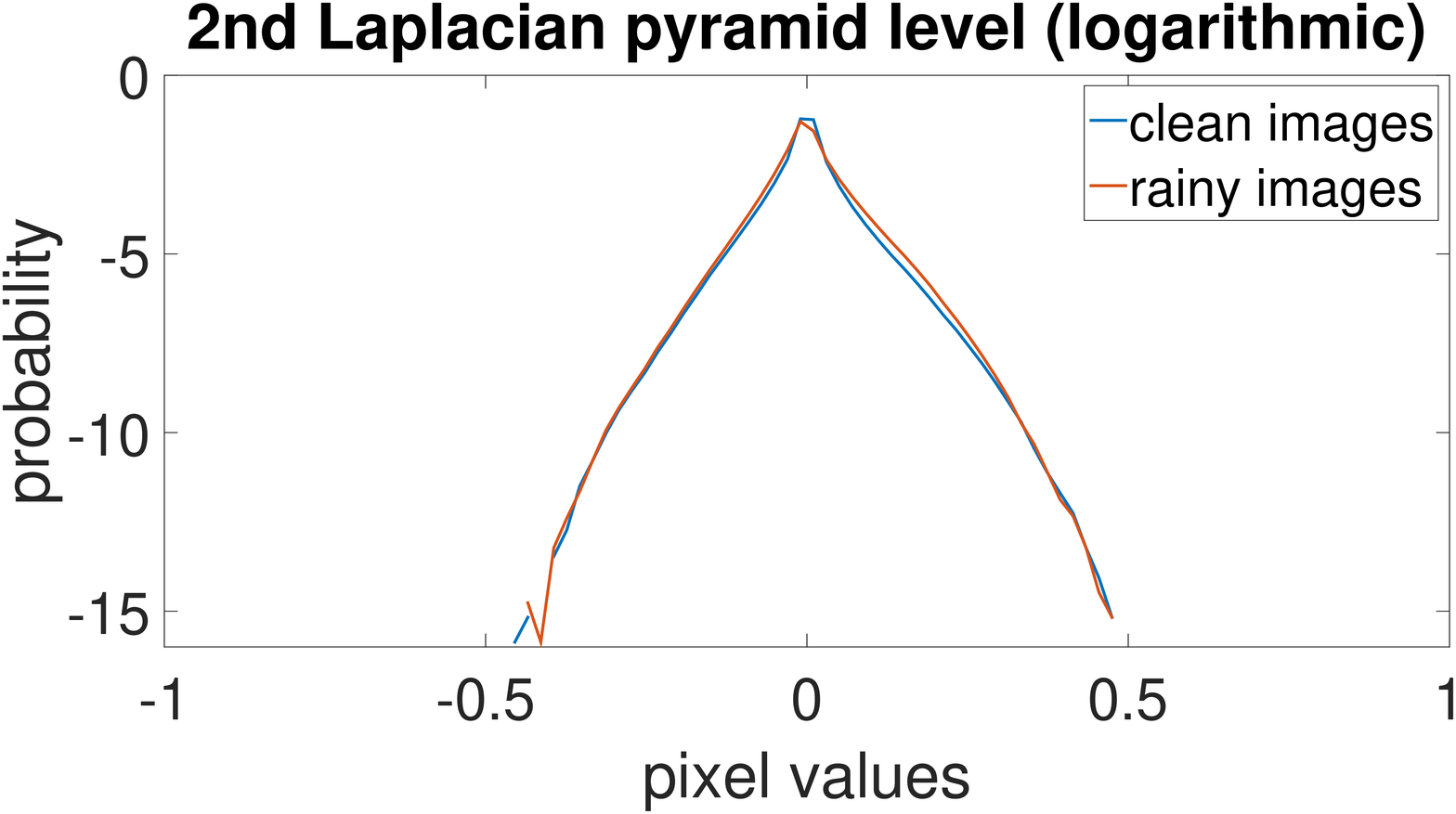}}
\subfigure[1st level]{\includegraphics[width = 2.1in]{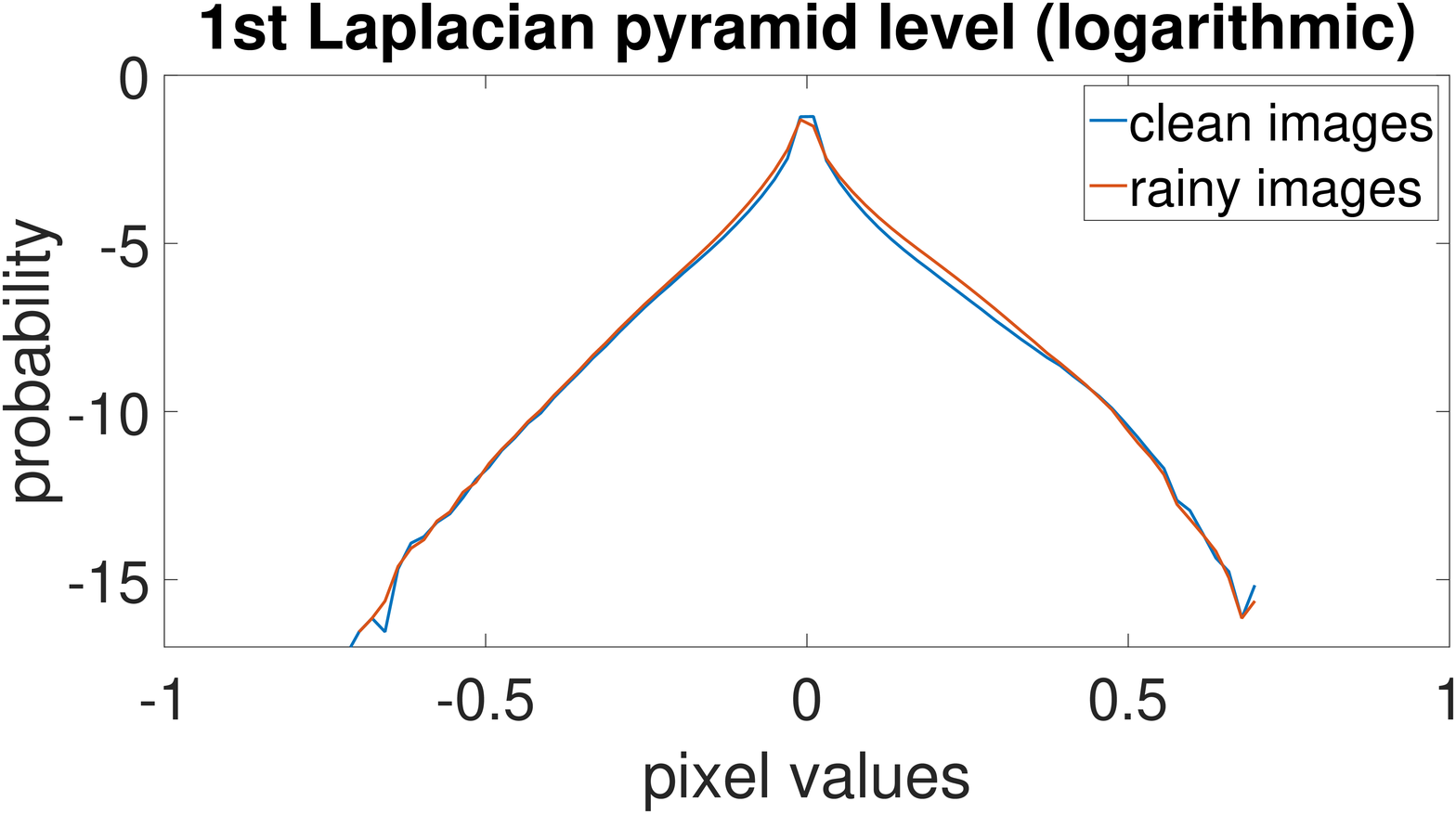}}
\caption{Statistical histogram distributions of 200 clean and rainy pairs from \cite{Yang2017Deep}. To highlight the tail error, (c)-(f) are logarithmic transformed.} \label{fig.hist2}
\end{center}
\end{figure*}
\subsection{Stage 2: Sub-network structure}
After decomposing $\bf{X}$ into different pyramid levels, we build a set of sub-networks independently for each level to predict a corresponding clean Gaussian pyramid $G(\bf{Y})$. All the sub-networks have the same network structure with different numbers of kernels. We adopt residual learning \cite{he2016deep} for each network structure and recursive blocks \cite{tai2017image} to reduce parameters. The sub-network structure can be expressed as follows:

\paragraph{Feature extraction} The first layer extracts features from the $n$th input level,
\begin{align}
\label{eq.Feature}
{\bf{H}}_{n,0} = \sigma({{\bf{W}}_n^0} * {{L_n}({\bf{X}})}+ {{\bf{b}}_n^0}),
\end{align}
where $\bf{H}$ indexes the feature map, $*$ is the convolution operation, $\bf{W}$ are weights and $\bf{b}$ are biases. $\sigma$ is an activation function for non-linearity.

\paragraph{Recursive block} To reduce the number of parameters, we build intermediate inference layers in a recursive fashion. The basic idea is to share parameters among recursive blocks. Motivated by our experiments, we adopt three convolutional operations in each recursive block. Calculations in the $t$th recursive block are
\begin{align}
\label{eq.Recursive}
{\bf{F}}_{n,t}^1 &= \sigma({{\bf{W}}_n^1} * {\bf{H}}_{n,t-1} + {{\bf{b}}_n^1}), \\
{\bf{F}}_{n,t}^2 &= \sigma({{\bf{W}}_n^2} * {\bf{F}}_{n,t}^1 + {{\bf{b}}_n^2}), \\
{\bf{F}}_{n,t}^3 &= {{\bf{W}}_n^3} * {\bf{F}}_{n,t}^2 + {{\bf{b}}_n^3},
\end{align}
where ${\bf{F}}^{\left\{{1,2,3}\right\}}$  are intermediate features in the recursive block, ${\bf{W}}^{\left\{{1,2,3}\right\}}$ and  ${\bf{b}}^{\left\{{1,2,3}\right\}}$ are shared parameters among $T$ recursive blocks and $t = 1,...,T$.

To help propagate information and back-propagate gradients, the output feature map ${\bf{H}}_{n,t}$ of the $t$th recursive block is calculated by adding ${\bf{H}}_{n,0}$:
\begin{align}
\label{eq.Recursive_out}
{\bf{H}}_{n,t} = \sigma({\bf{F}}_{n,t}^3 + {\bf{H}}_{n,0}).
\end{align}

\paragraph{Gaussian pyramid reconstruction} To obtain the output level of the pyramid, the reconstruction layer is expressed as:
\begin{align}
\label{eq.net_out}
{{L_n}({\bf{Y}})} = ({{\bf{W}}_n^4} * {\bf{H}}_{n,T} + {{\bf{b}}_n^4}) + {{L_n}({\bf{X}})}.
\end{align}
After obtaining the output of the Laplacian pyramid ${L}({\bf{Y}})$, the corresponding Gaussian pyramid of the derained image can be reconstructed by
\begin{equation}
\begin{split}
\label{eq.final_out}
{{G_N}({\bf{Y}})} &= max(0, {{L_N}({\bf{Y}})}), \\
{{G_n}({\bf{Y}})} &= max(0, {L_{n}}({\bf{Y}}) + \text{upsample}({G_{n+1}}({\bf{Y}}))),
\end{split}
\end{equation}
where $n = 1,...,N-1$. Since each level of a Gaussian pyramid should equal or lager than 0, we use $x=max(0,x)$, which is actually the rectified linear units (ReLU) operation \cite{krizhevsky2012imagenet}, to simply correct the outputs. The final derained image is the bottom level of the Gaussian pyramid, i.e., ${G_1}(\bf{Y})$.

In methods \cite{Denton2015Deep,Lai2017LapSRN,Shen2017pyramid}, the authors build similar networks based on the image pyramid, which are the most related to our own work. However, these papers apply similar structures to other tasks such a image generation or super-resolution using different network approaches on the pyramid.

\subsection{Loss Function}
Given a training set $\{ {{\bf{X}}^{i}},{\bf{Y}}_{GT}^{i}\}_{i = 1}^M$, where $M$ is the number of training data and ${\bf{Y}}_{GT}$ is the ground truth, the most widely used loss function for training a network is mean squared error (MSE). However, MSE usually generates over-smoothed results due to the squared penalty that works poorly at edges in an image. Thus, for each sub-network we adopt different loss functions and minimize their combination. Following \cite{Zhao2017Loss}, we choose $\ell_1$ and SSIM \cite{Wang2014SSIM} as our loss functions. Specifically, as shown in Figure \ref{fig.hist}, since finer details and rain streaks exist in lower pyramid levels we use SSIM loss to train the corresponding sub-networks for better preserving high-frequency information. On the contrary, larger structures and smooth background areas exist in higher pyramid levels. Thus we use the $\ell_1$ loss to update the corresponding network parameters there. The overall loss function is
\begin{equation}
\begin{split}
\footnotesize
\label{eq.overall}
\mathcal{L}= &\frac{1}{M}\sum\limits_{i = 1}^M \{\sum\limits_{n = 1}^N{{\mathcal{L}^{\ell_1}({G_n}({{\bf{Y}}^i}),{G_n}({\bf{Y}}_{GT}^i}))} \\
&+ {\sum\limits_{n = 1}^2 {\mathcal{L}^{SSIM}({G_n}({{\bf{Y}}^i}),{G_n}({\bf{Y}}_{GT}^i)} )}\},
\end{split}
\end{equation}
where $\mathcal{L}^{SSIM}$ is the SSIM loss and  $\mathcal{L}^{\ell_1}$ is the $\ell_1$ loss. In this paper, we set the pyramid level $N=5$ based on our experiments. We use SSIM loss for levels ${\left\{{1,2}\right\}}$ and $\ell_1$ loss for all levels.

\subsection{Removing batch normalization}
As one of the most effective way to alleviate the internal co-variate shift, batch normalization (BN) \cite{Ioffe2015Batch} is widely adopted before the nonlinearity in each layer in existing deep learning based methods. However, we argue that by introducing image pyramid technology, BN can be removed to improve the flexibility of networks. This is because BN constrains the feature maps to obey a Gaussian distribution. While during our experiments, we found that distributions of lower Laplacian pyramid levels of both clean and rainy images are sparse. To demonstrate this viewpoint, in Figure \ref{fig.hist2}, we show the histogram distributions of each Laplacian pyramid level from 200 clean and light rainy training image pairs from \cite{Yang2017Deep}. As can be seen, compared to the image domain in Figure \ref{fig.hist2}(a), distributions of lower pyramid levels, i.e., Figures \ref{fig.hist2}(c) to (f), are more sparse and do not obey Gaussian distribution. This implies that we do not need BN to further constrain the feature maps since the mapping problem already becomes easy to handle. Moreover, removing BN can sufficiently reduce GPU memory usage since the BN layers consume the same amount of memory as the preceding convolutional layers. Based on the above observation and analysis, we remove BN layers from our network to improve flexibility and reduce parameter numbers and computing resource.

\subsection{Parameter settings}
We decompose an RGB image into a $5$-level Laplacian pyramid by using a fixed smoothing kernel $[0.0625, 0.25, 0.375, 0.25, 0.0625]$, which is also used to reconstruct the Gaussian pyramid. In our network architecture, each sub-network has the same structure with a different numbers of kernels. The kernel sizes for ${\bf{W}}^{\left\{{0,1,3,4}\right\}}$ are $3\times 3$. For  ${\bf{W}}^{\left\{{2}\right\}}$, the kernel size is $1\times 1$ to further increase non-linearity and reduce parameters. The number of recursive blocks is $T = 5$ for each sub-network. For the activation function $\sigma$, we use the leaky rectified linear units (LReLUs) \cite{Maas2013Rectifier} with a negative slope of 0.2.

\begin{figure*}
\centering
\includegraphics[width = 6.5in]{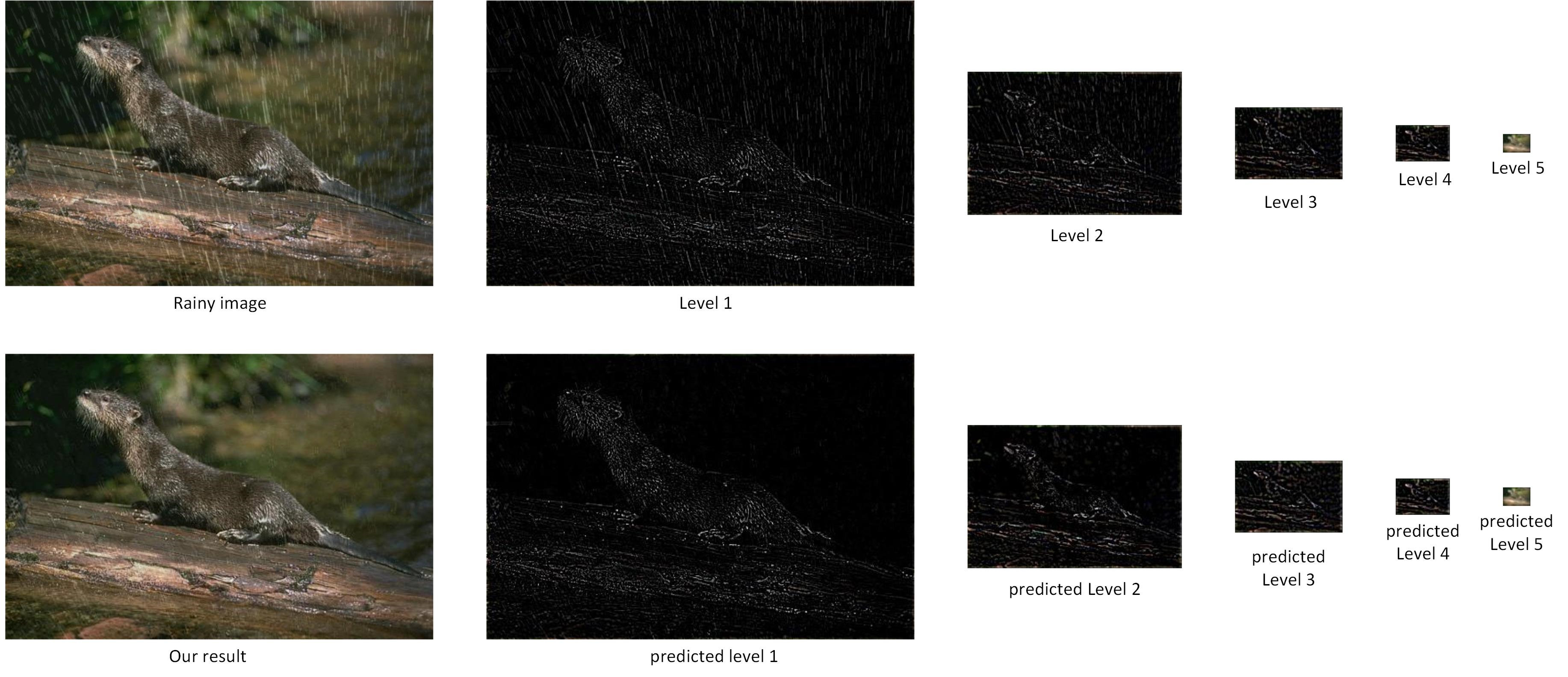}
\caption{One example of intermediate results predicted by our LPNet.} \label{fig.intermediate}
\end{figure*}

Moreover, as shown in the last row of Figure \ref{fig.hist}, higher levels are closer to an identity mapping since rain streaks only remain in lower levels. This means for higher levels, fewer parameters are required for learning a good network. Thus, from low to high levels, we set the kernel numbers to $16, 8, 4, 2$ and $1$, respectively. Since the top level is a tiny and smoothed version of image and rain streaks remain in high-frequency parts, the function of top level sub-network is more like a simple global contrast adjustment. Thus we set the kernel numbers to $1$ kernel for the top level. As shown in Figure \ref{fig.flowchart}, by connecting the up-sampled version of the output from the higher level, the direct prediction of all sub-networks is actually the clean Laplacian pyramid. We show the intermediate results predicted by each sub-network in Figure \ref{fig.intermediate}. It is clear that rain streaks remain in lower levels while higher levels are almost the same. This demonstrates that our diminishing parameter setting is reasonable. As a result, the total number of trainable parameters is only $7,548$, far fewer than the hundreds of thousands often encountered in deep learning.

\subsection{Training details}
We use synthetic rainy images from \cite{Yang2017Deep} as our training data. This dataset contains 1800 images with heavy rain and 200 images with light rain. We randomly generate one million $80 \times 80$ clean/rainy patch pairs. We use TensorFlow \cite{abadi2016tensorflow} to train LPNet using the Adam solver \cite{Kingma2014Adam} with a mini-batch size of $10$. We set the learning rate as $0.001$ and finish the training after $3$ epochs. The whole network is trained in a end-to-end fashion.
\begin{figure*}[!htp]
\centering
\includegraphics[width = .13\textwidth]{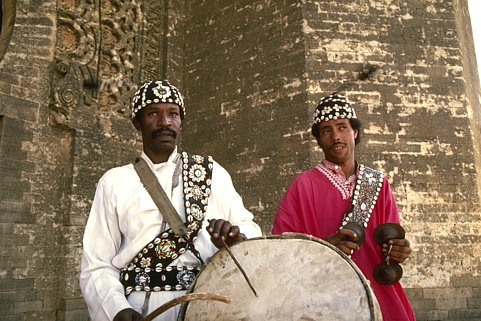}
\includegraphics[width = .13\textwidth]{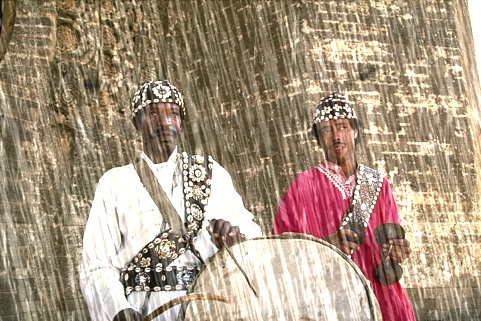}
\includegraphics[width = .13\textwidth]{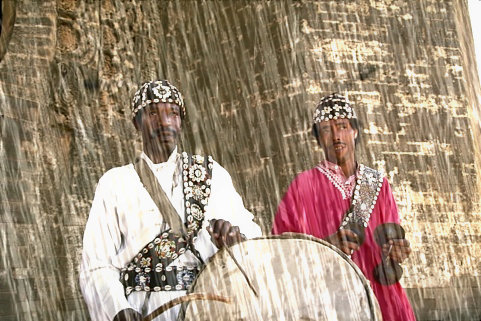}
\includegraphics[width = .13\textwidth]{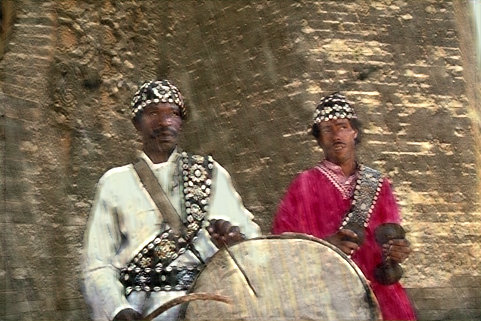}
\includegraphics[width = .13\textwidth]{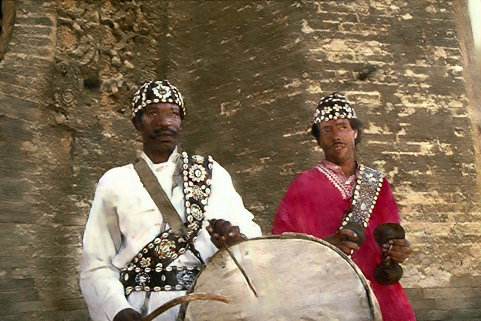}
\includegraphics[width = .13\textwidth]{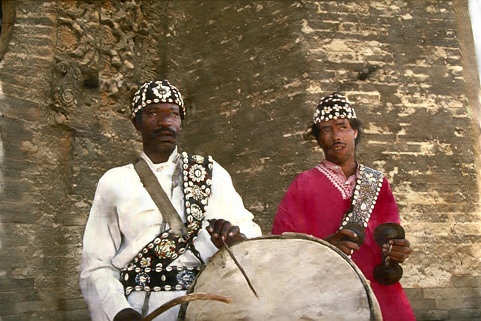}
\includegraphics[width = .13\textwidth]{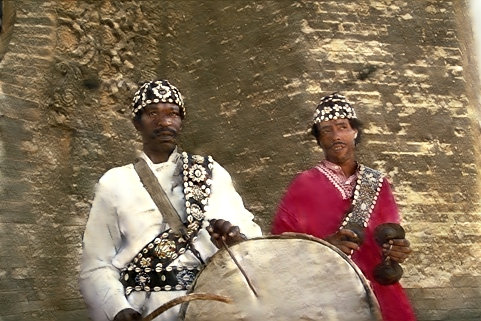}  \\
\subfigure[Ground Truth]{\includegraphics[width = .13\textwidth]{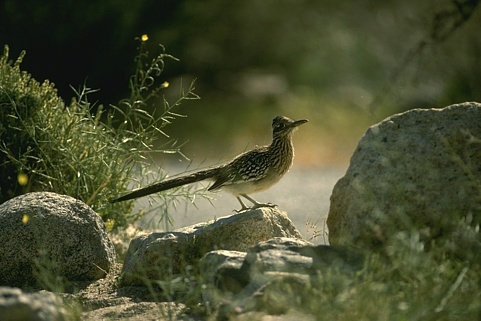}}
\subfigure[Rainy images]{\includegraphics[width = .13\textwidth]{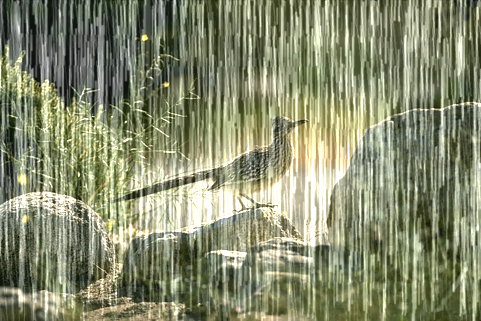}}
\subfigure[GMM]{\includegraphics[width = .13\textwidth]{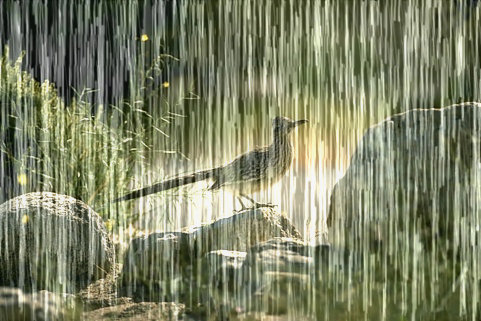}}
\subfigure[SRCNN]{\includegraphics[width = .13\textwidth]{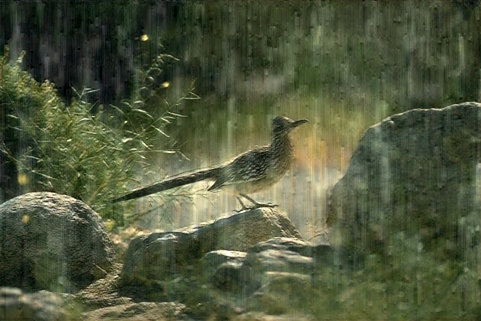}}
\subfigure[DDN]{\includegraphics[width = .13\textwidth]{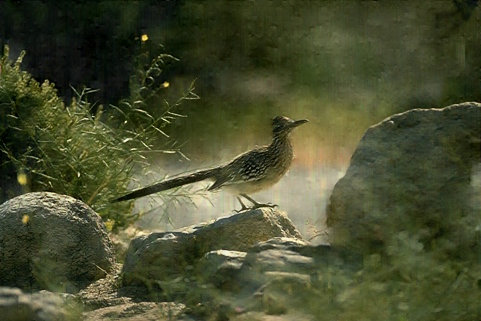}}
\subfigure[JORDER]{\includegraphics[width = .13\textwidth]{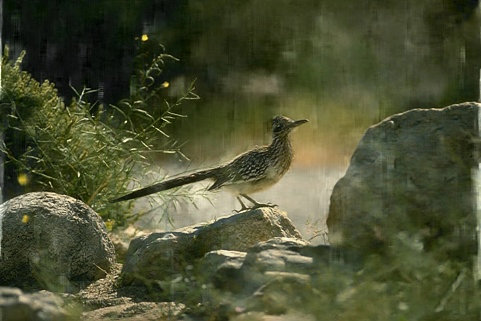}}
\subfigure[Our LPNet]{\includegraphics[width = .13\textwidth]{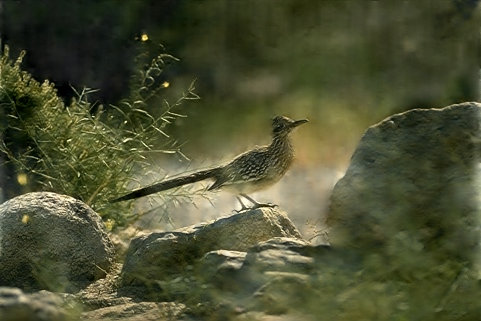}}
\caption{Two synthetic images from ``\emph{Rain100H}'' \cite{Yang2017Deep} with different rain orientations and magnitudes.} \label{fig.syntheticH}
\end{figure*}
\begin{figure*}
\centering
\includegraphics[width = .13\textwidth]{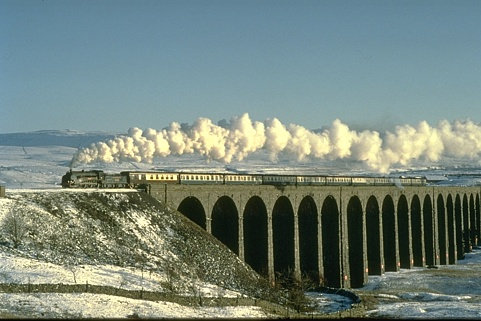}
\includegraphics[width = .13\textwidth]{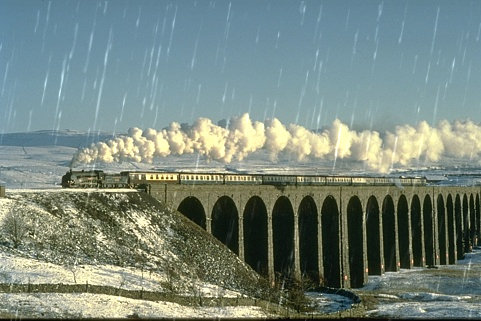}
\includegraphics[width = .13\textwidth]{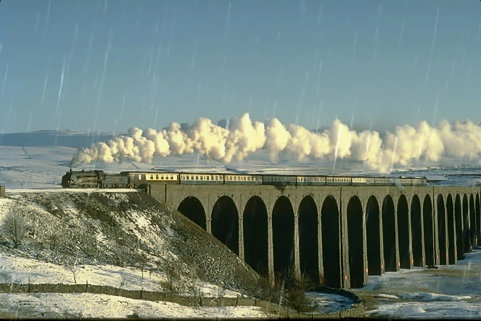}
\includegraphics[width = .13\textwidth]{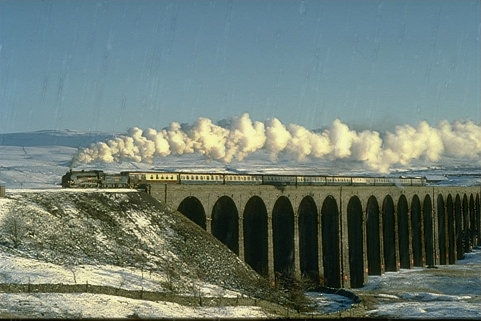}
\includegraphics[width = .13\textwidth]{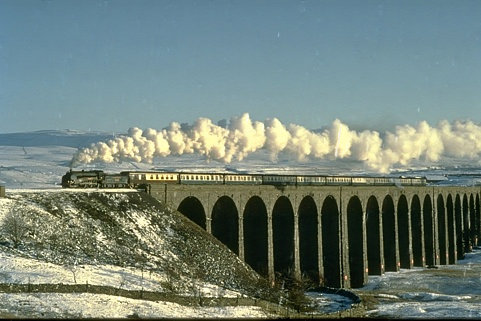}
\includegraphics[width = .13\textwidth]{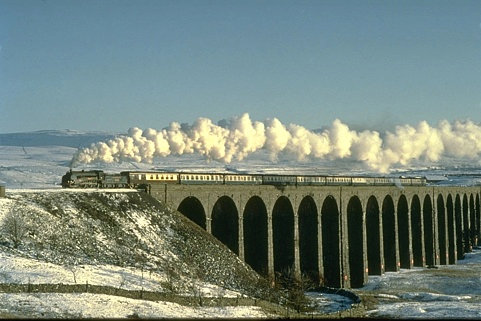}
\includegraphics[width = .13\textwidth]{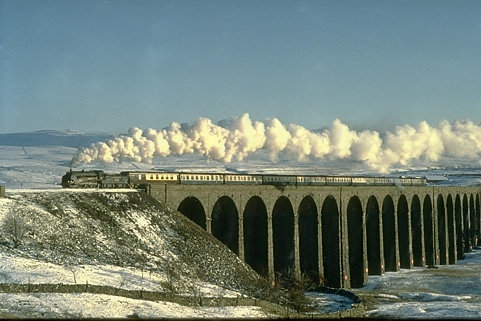}  \\
\subfigure[Ground Truth]{\includegraphics[width = .13\textwidth]{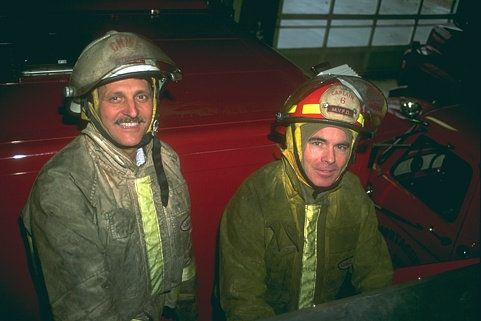}}
\subfigure[Rainy images]{\includegraphics[width = .13\textwidth]{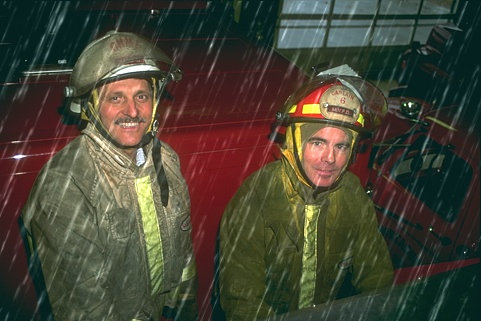}}
\subfigure[GMM]{\includegraphics[width = .13\textwidth]{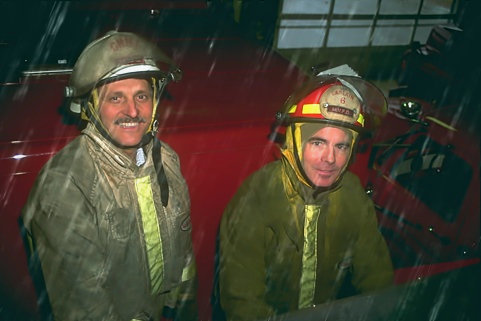}}
\subfigure[SRCNN]{\includegraphics[width = .13\textwidth]{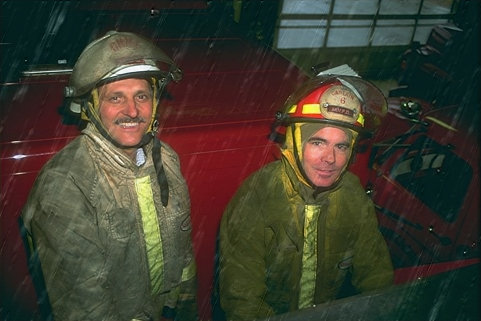}}
\subfigure[DDN]{\includegraphics[width = .13\textwidth]{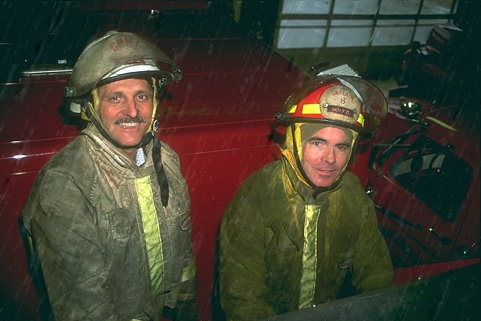}}
\subfigure[JORDER]{\includegraphics[width = .13\textwidth]{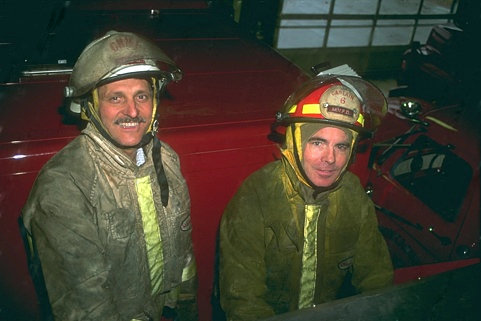}}
\subfigure[Our LPNet]{\includegraphics[width = .13\textwidth]{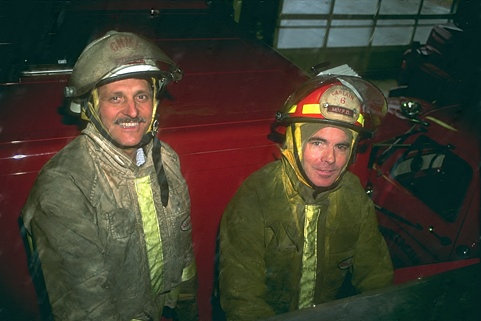}}
\caption{Two synthetic images from ``\emph{Rain100L}'' \cite{Yang2017Deep}  with different rain orientations and magnitudes.} \label{fig.syntheticL}
\end{figure*}
\begin{figure*}
\centering
\includegraphics[width = .13\textwidth]{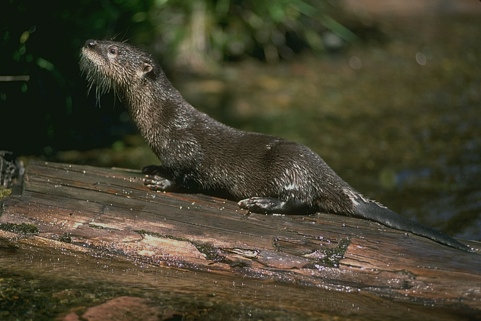}
\includegraphics[width = .13\textwidth]{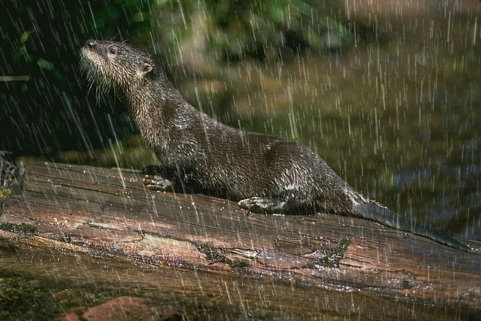}
\includegraphics[width = .13\textwidth]{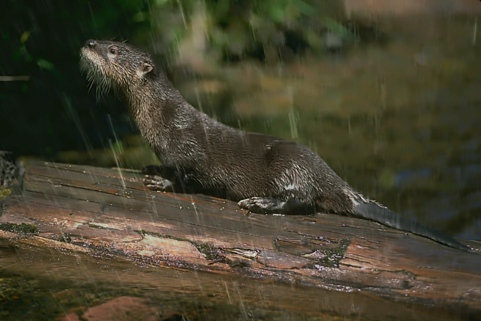}
\includegraphics[width = .13\textwidth]{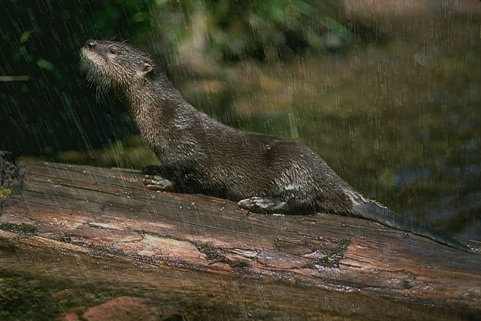}
\includegraphics[width = .13\textwidth]{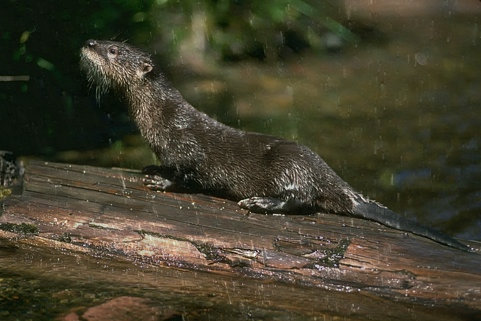}
\includegraphics[width = .13\textwidth]{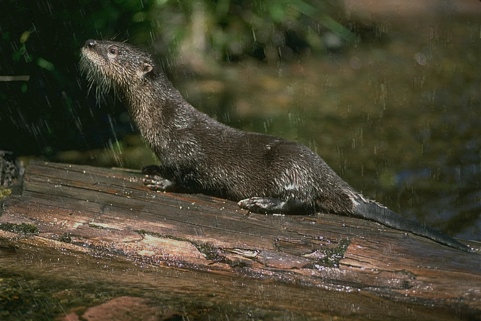}
\includegraphics[width = .13\textwidth]{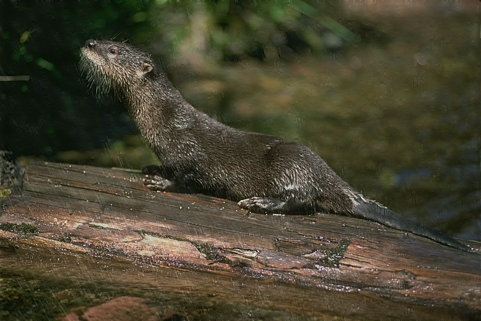}  \\
\subfigure[Ground Truth]{\includegraphics[width = .13\textwidth]{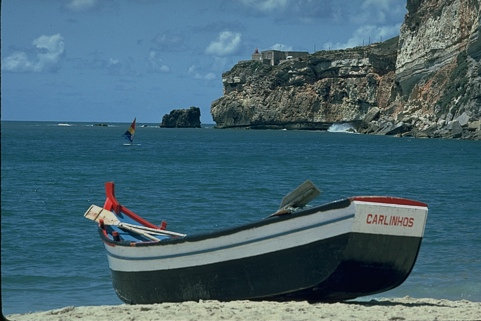}}
\subfigure[Rainy images]{\includegraphics[width = .13\textwidth]{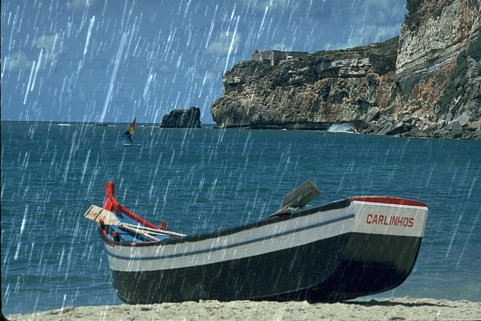}}
\subfigure[GMM]{\includegraphics[width = .13\textwidth]{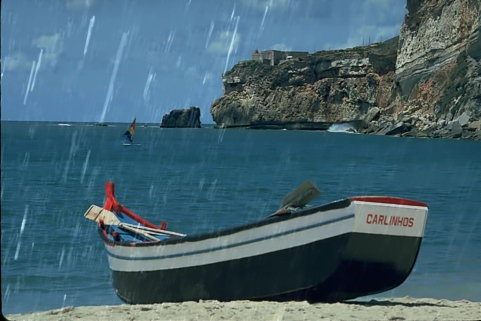}}
\subfigure[SRCNN]{\includegraphics[width = .13\textwidth]{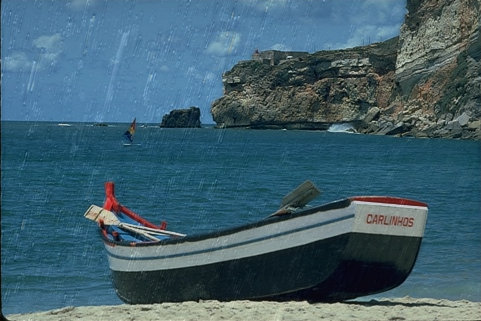}}
\subfigure[DDN]{\includegraphics[width = .13\textwidth]{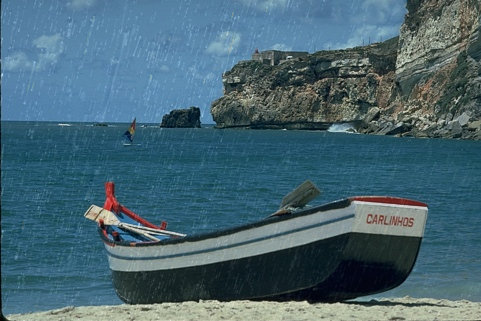}}
\subfigure[JORDER]{\includegraphics[width = .13\textwidth]{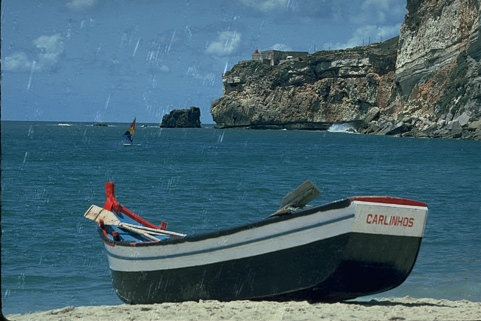}}
\subfigure[Our LPNet]{\includegraphics[width = .13\textwidth]{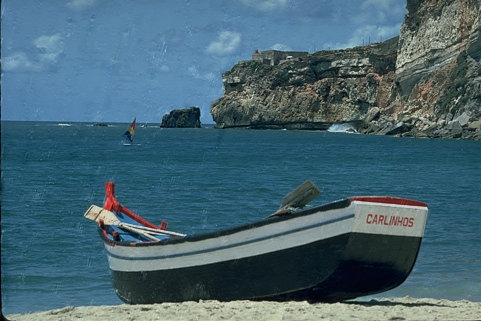}}
\caption{Two synthetic images from ``\emph{Rain12}'' \cite{Li2016Rain} with different rain orientations and magnitudes.} \label{fig.synthetic12}
\end{figure*}

\section{Experiments}
We compare our LPNet with four state-of-the-art deraining methods: the Gaussian Mixture Model (GMM) of \cite{Li2016Rain}, a CNN baseline SRCNN \cite{dong2016image}, the deep detail network (DDN) of \cite{fu2017removing} and joint rain detection and removal (JORDER) \cite{Yang2017Deep}, which is also a deep learning method. For fair comparison, all CNN based methods are retrained on the same training dataset.

\subsection{Synthetic data}
Three synthetic datasets are chosen for comparison. Two of them are from \cite{Yang2017Deep} and each one contains $100$ images. One is synthesized with heavy rain called \emph{Rain100H} and the other one is with light rain called \emph{Rain100L}. The third dataset called  \emph{Rain12} is from \cite{Li2016Rain} which contains $12$ synthetic images. All testing results shown are not included in the training data. Following \cite{Yang2017Deep}, for each CNN method we train two models, one for heavy and for light rain datasets. The model trained on the light rainy dataset is used to test \emph{Rain12}.

Figures \ref{fig.syntheticH} to \ref{fig.synthetic12} shows visual results from each dataset. As can be seen, GMM \cite{Li2016Rain} fails to remove rain streaks form heavy rainy images. SRCNN \cite{dong2016image} and DDN \cite{fu2017removing} are able to remove the rain streaks while tend to generate obvious artifacts. Our LPNet has comparable visual results with JORDER and outperforms other methods.

We also adopt PSNR and SSIM  \cite{Wang2014SSIM} to perform quantitative evaluations in Table \ref{tab.SSIM}. Our method has comparable SSIM values with JORDER while outperforming other methods, in agreement with the visual results. Though our result has a lower PSNR value than JORDER method, the visual quality is comparable. This is because PSNR is calculated based on the mean squared error (MSE), which measures global pixel errors without considering local image characters. Moreover, as shown in Table \ref{tab.SSIM} our LPNet contains far fewer parameters, potentially making LPNet more suitable for storage, e.g., in mobile devices.
\begin{table*}
\centering
\caption{Average SSIM and PSNR values on synthesized images.}
\begin{tabular}{|*{13}{c|}}
\hline
& \multicolumn{2}{|c|}{Rainy images} & \multicolumn{2}{|c|}{GMM  \cite{Li2016Rain}} & \multicolumn{2}{|c|}{SRCNN \cite{dong2016image}}& \multicolumn{2}{|c|}{DDN \cite{fu2017removing}} & \multicolumn{2}{|c|}{JORDER \cite{Yang2017Deep}}& \multicolumn{2}{|c|}{Our LPNet} \\\cline{1-13}
& SSIM &  PSNR  & SSIM    & PSNR & SSIM    & PSNR   & SSIM &  PSNR  & SSIM    & PSNR & SSIM    & PSNR\\
 \hline
\emph{Rain100H}   & 0.38&13.56&  0.43&15.05 &    0.70&22.84  &   0.76&21.92   &   \textbf{0.83}&\textbf{26.54}  &   0.81&23.73 \\
\hline
\emph{Rain100L}   & 0.84&26.90&  0.86&28.65 &    0.91&29.39  &   0.93&32.16   &   \textbf{0.97}&\textbf{36.63}  &   0.95&34.26 \\
\hline
\emph{Rain12}     & 0.86&30.14&  0.91&32.02 &    0.92&31.90  &   0.94& 31.76   &   \textbf{0.95}&33.92  &   \textbf{0.95}&\textbf{35.35} \\
\hline
Parameters \# &\multicolumn{2}{|c|}{-} & \multicolumn{2}{|c|}{-} & \multicolumn{2}{|c|}{20,099}&\multicolumn{2}{|c|}{57,369} & \multicolumn{2}{|c|}{369,792} & \multicolumn{2}{|c|}{\textbf{7,548}} \\\cline{1-13}
\end{tabular}
\label{tab.SSIM}
\end{table*}

\subsection{Real-world data}
\begin{figure*}
\centering
\includegraphics[width = .16\textwidth]{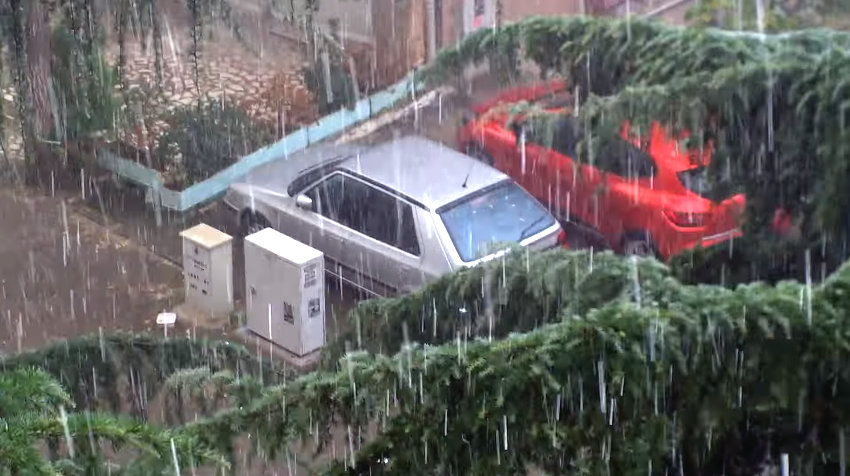}
\includegraphics[width = .16\textwidth]{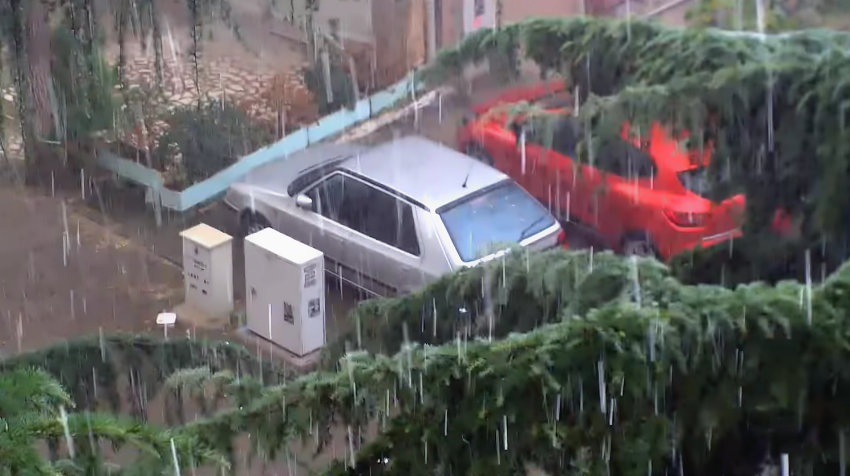}
\includegraphics[width = .16\textwidth]{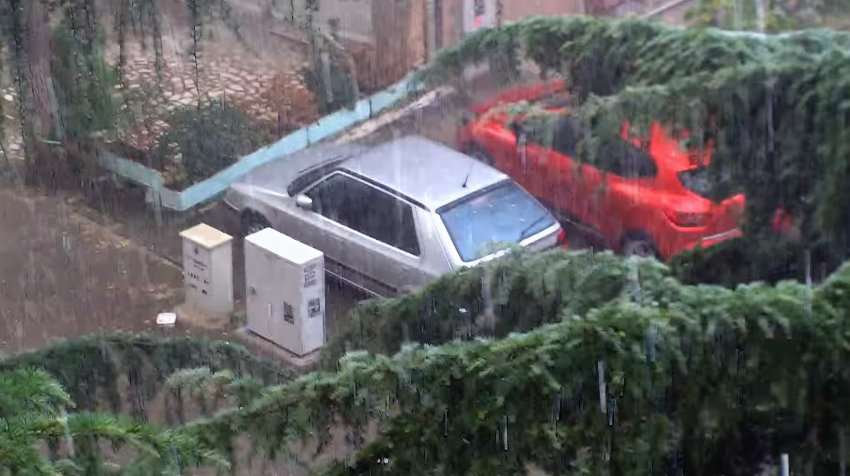}
\includegraphics[width = .16\textwidth]{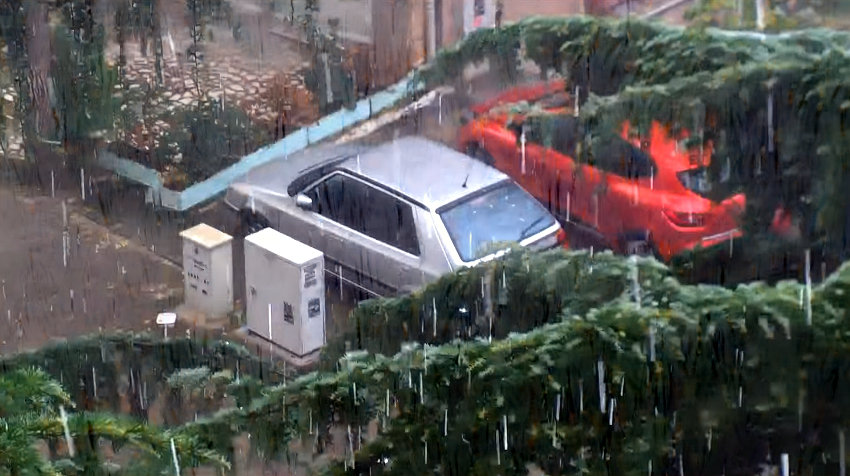}
\includegraphics[width = .16\textwidth]{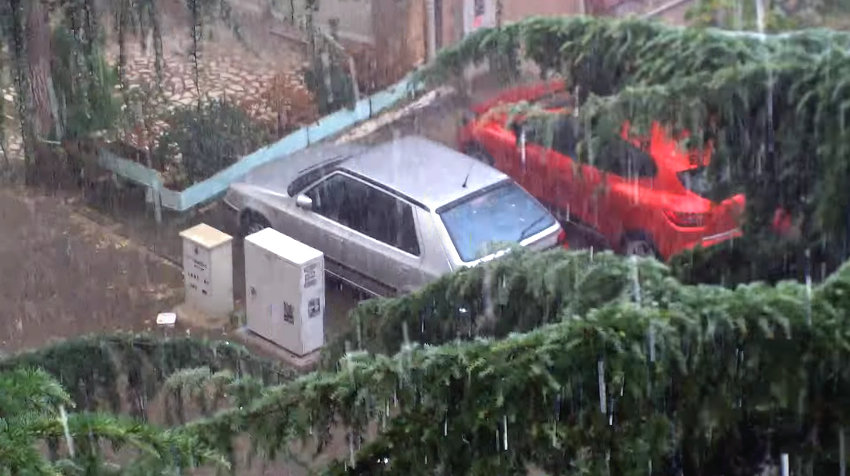}
\includegraphics[width = .16\textwidth]{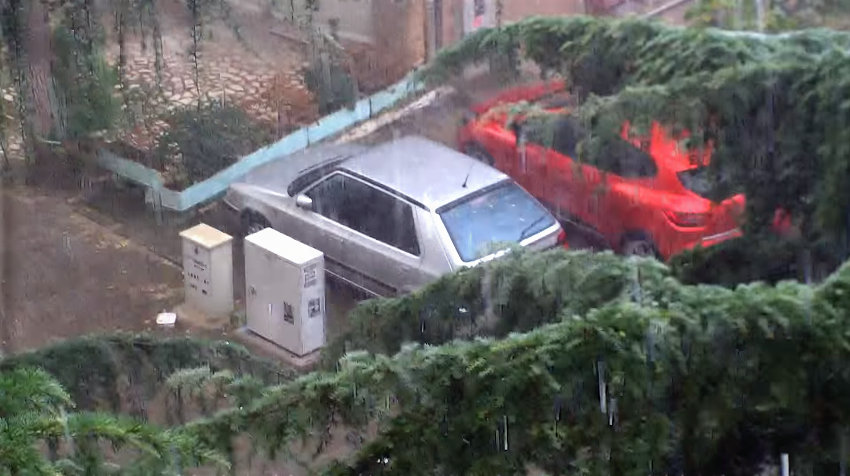}  \\ \vspace{0.07in}
\includegraphics[width = .16\textwidth]{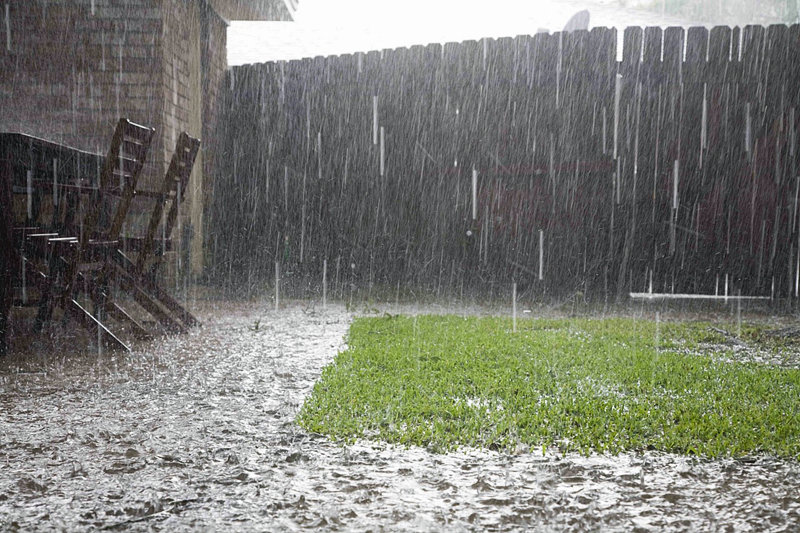}
\includegraphics[width = .16\textwidth]{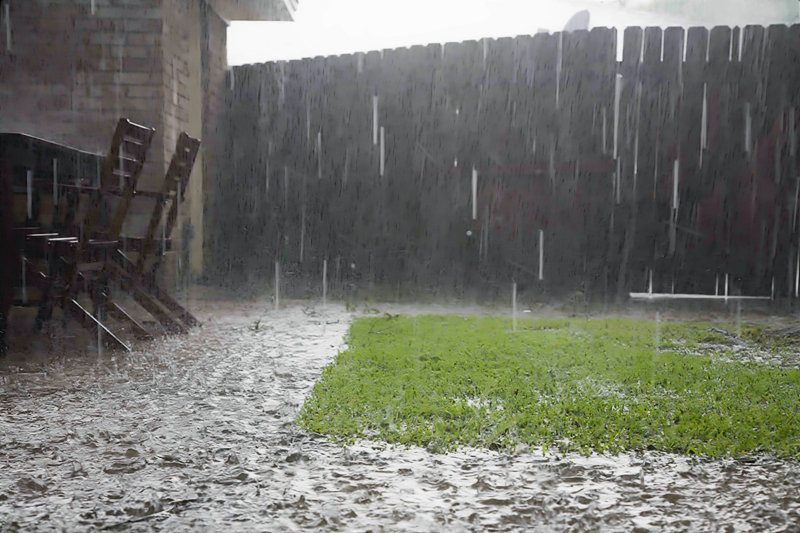}
\includegraphics[width = .16\textwidth]{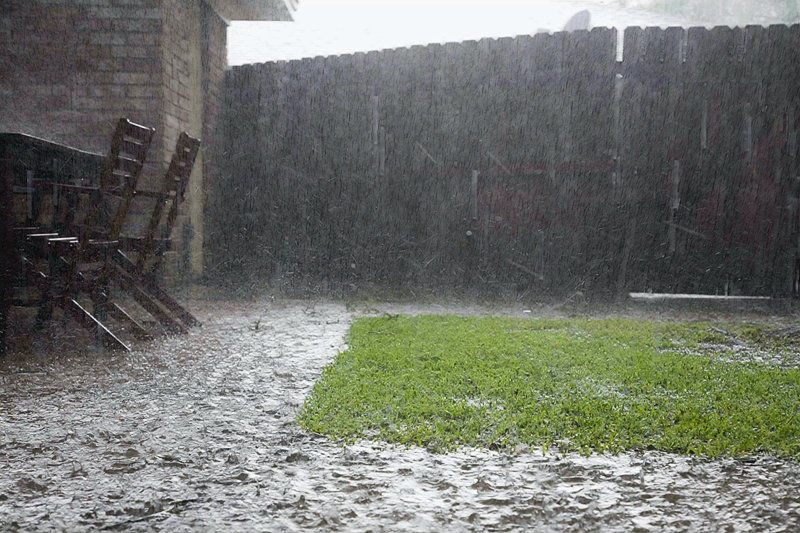}
\includegraphics[width = .16\textwidth]{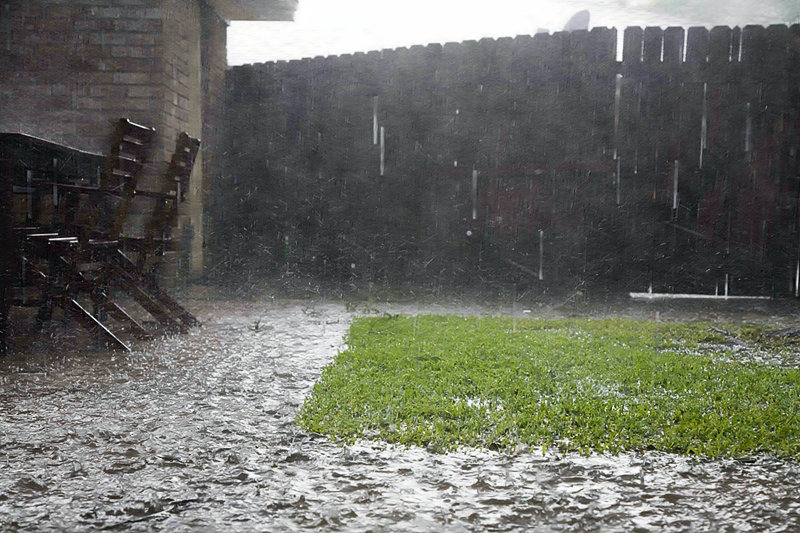}
\includegraphics[width = .16\textwidth]{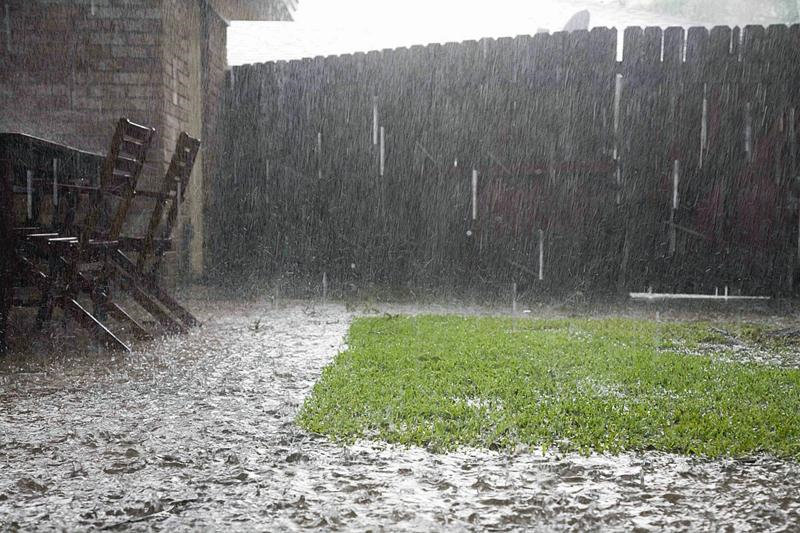}
\includegraphics[width = .16\textwidth]{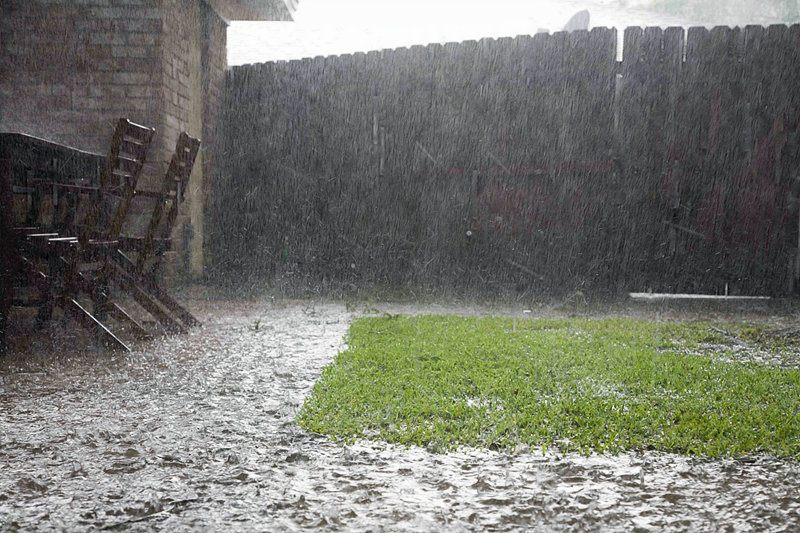} \\ \vspace{0.07in}
\includegraphics[width = .16\textwidth]{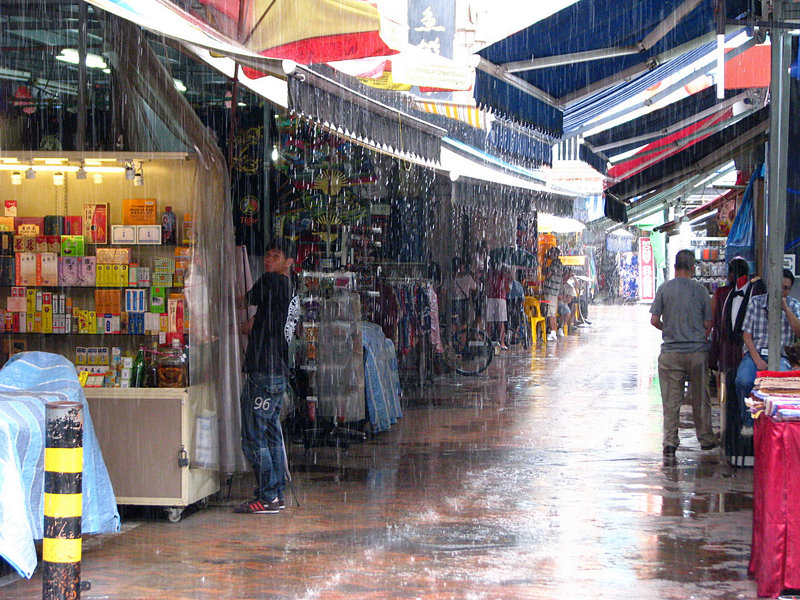}
\includegraphics[width = .16\textwidth]{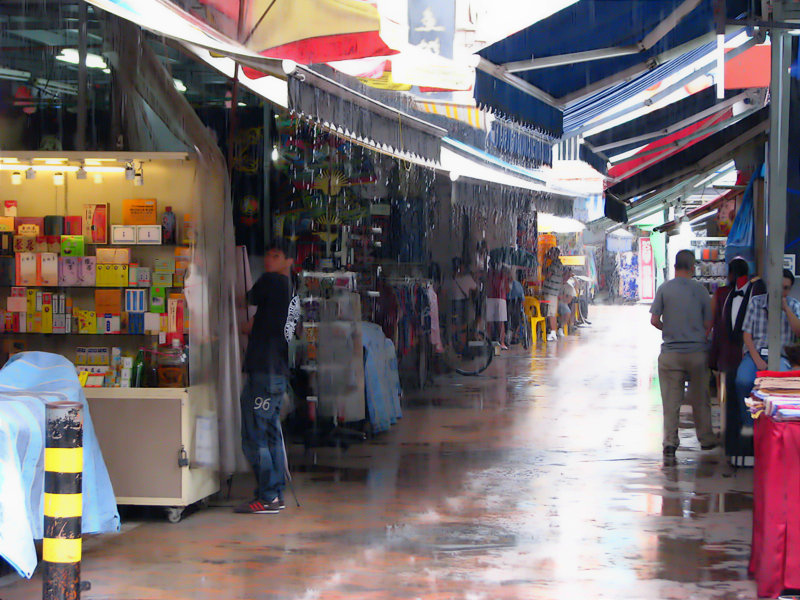}
\includegraphics[width = .16\textwidth]{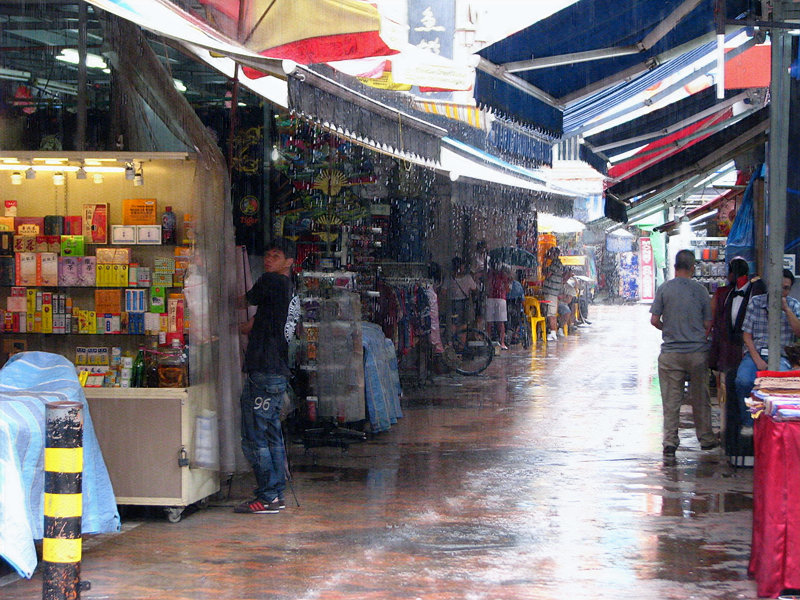}
\includegraphics[width = .16\textwidth]{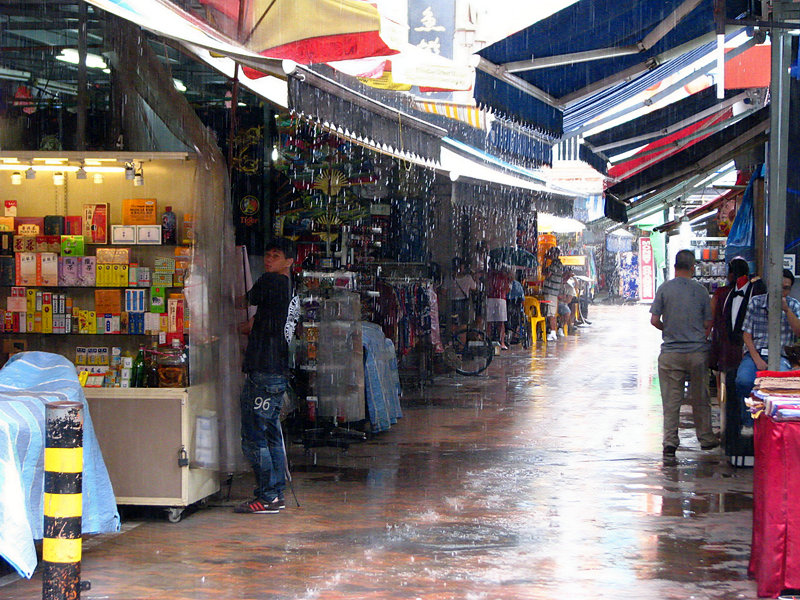}
\includegraphics[width = .16\textwidth]{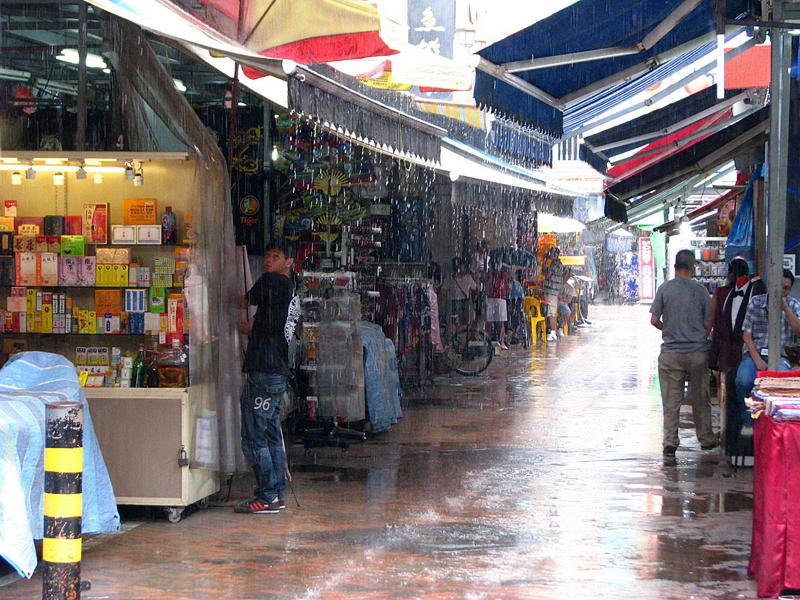}
\includegraphics[width = .16\textwidth]{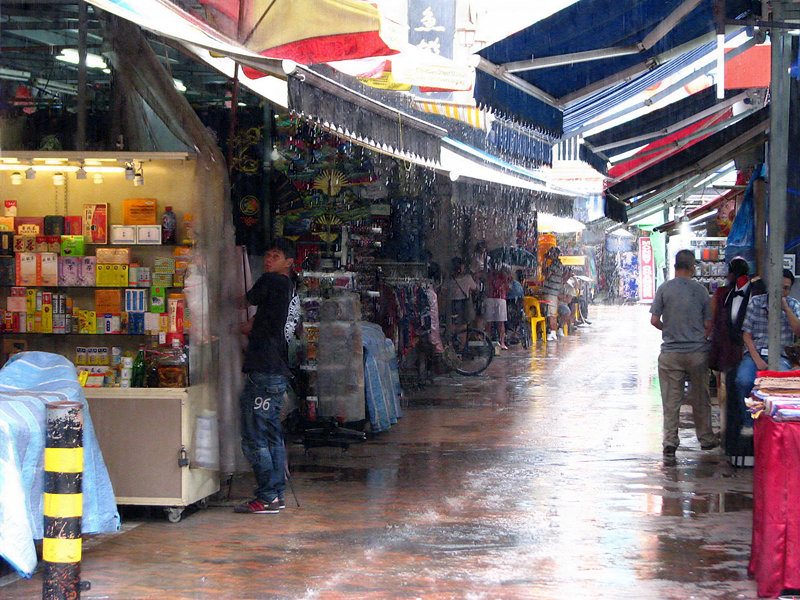} \\ \vspace{0.07in}
\includegraphics[width = .16\textwidth]{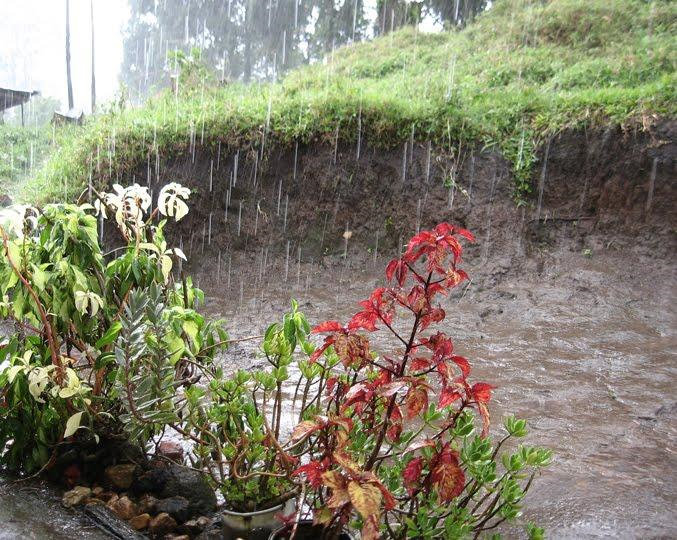}
\includegraphics[width = .16\textwidth]{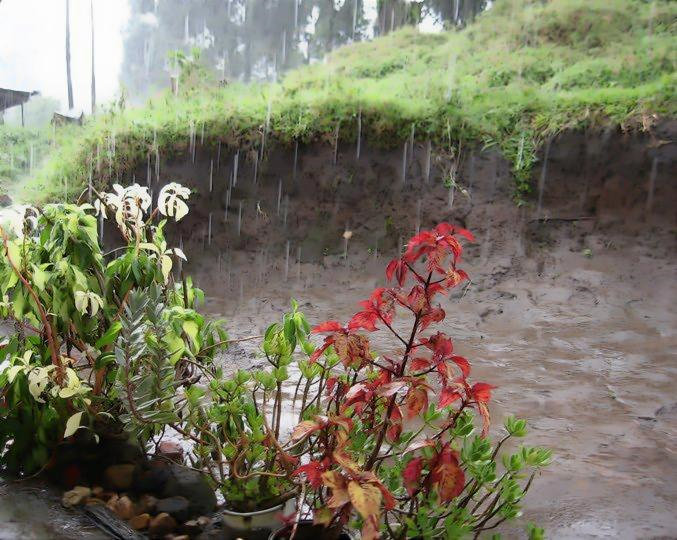}
\includegraphics[width = .16\textwidth]{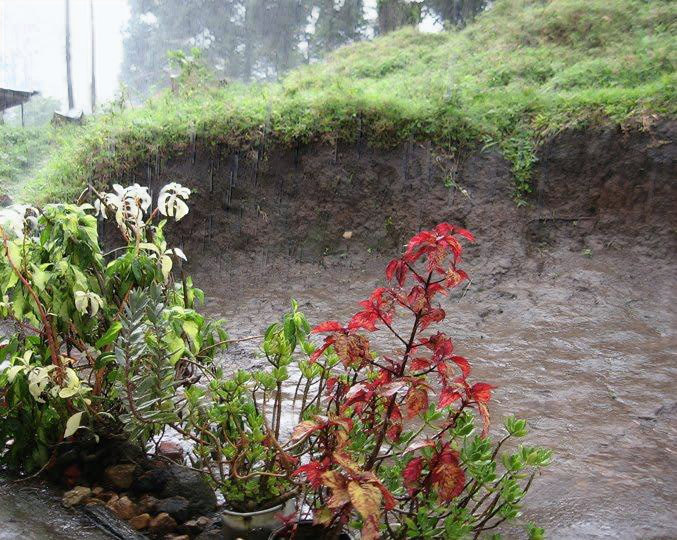}
\includegraphics[width = .16\textwidth]{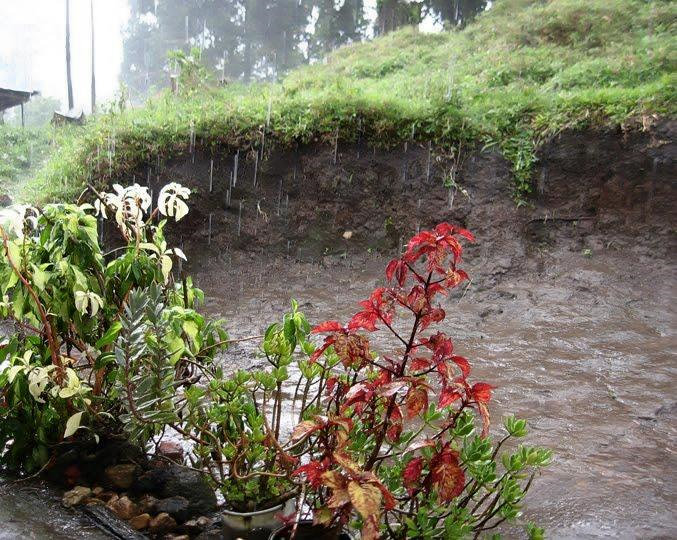}
\includegraphics[width = .16\textwidth]{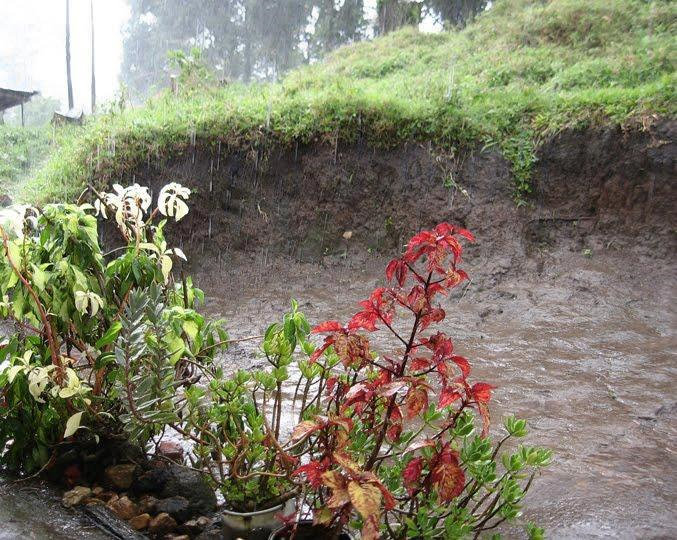}
\includegraphics[width = .16\textwidth]{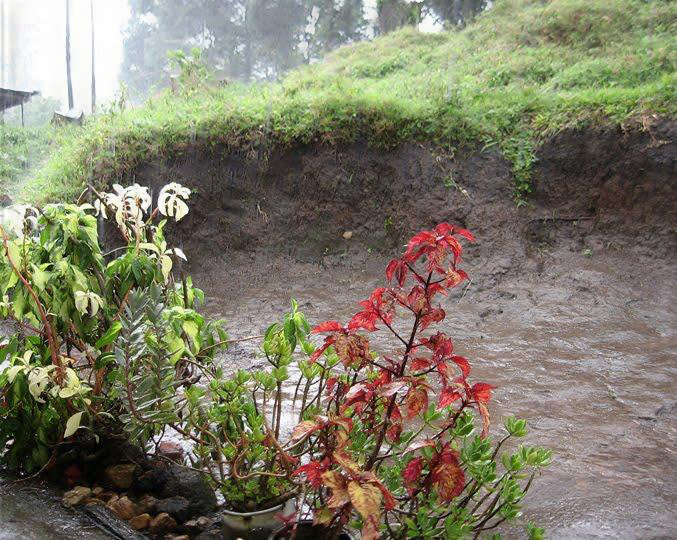} \\
\subfigure[Rainy images]{\includegraphics[width = .16\textwidth]{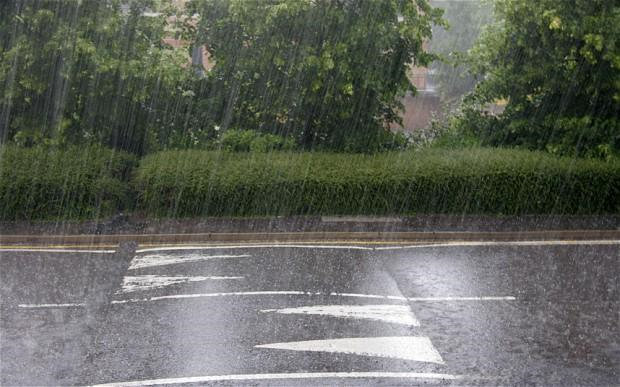}}
\subfigure[{GMM}]{\includegraphics[width = .16\textwidth]{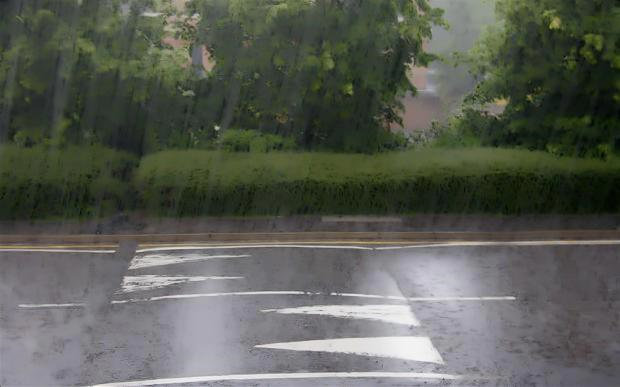}}
\subfigure[{SRCNN}]{\includegraphics[width = .16\textwidth]{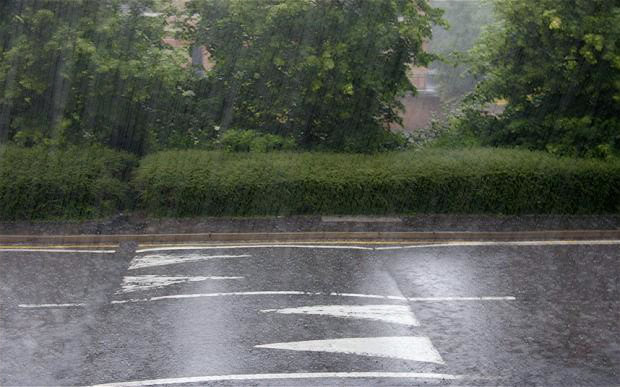}}
\subfigure[{DDN}]{\includegraphics[width = .16\textwidth]{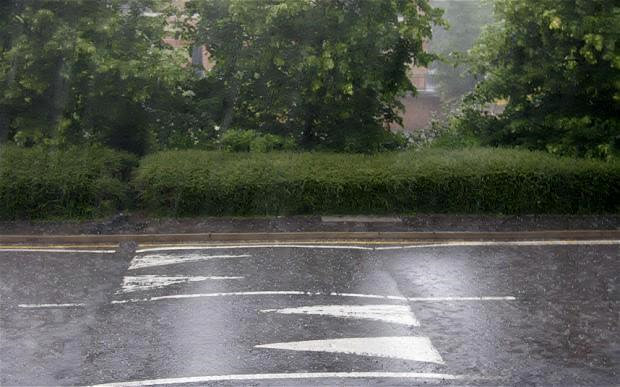}}
\subfigure[{JORDER}]{\includegraphics[width = .16\textwidth]{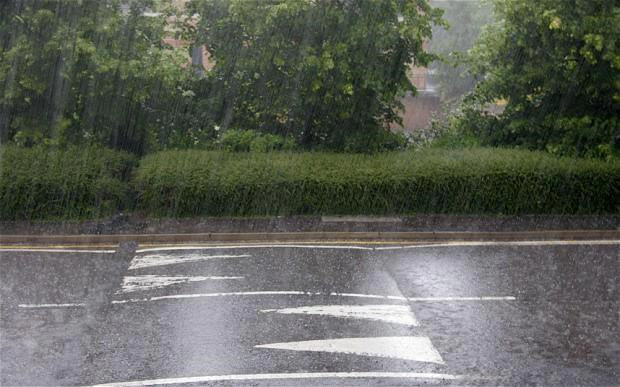}}
\subfigure[{Our LPNet}]{\includegraphics[width = .16\textwidth]{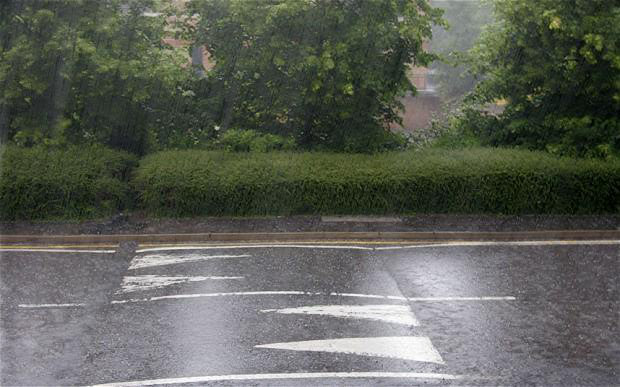}}
\caption{Three results on real-world rainy images with different rain orientations and magnitudes.} \label{fig.real}
\end{figure*}

In this section, we show that the LPNet learned on synthetic training data still performs well on real-world data. Figure \ref{fig.real} shows five visual results on real-world images. The model trained on the dataset with light rain is used for testing on real-world images. As can be seen, LPNet generates consistently promising derained results on images with different kinds of rain streaks.

\begin{figure}[!htp]
\centering
\includegraphics[width = 3.5in]{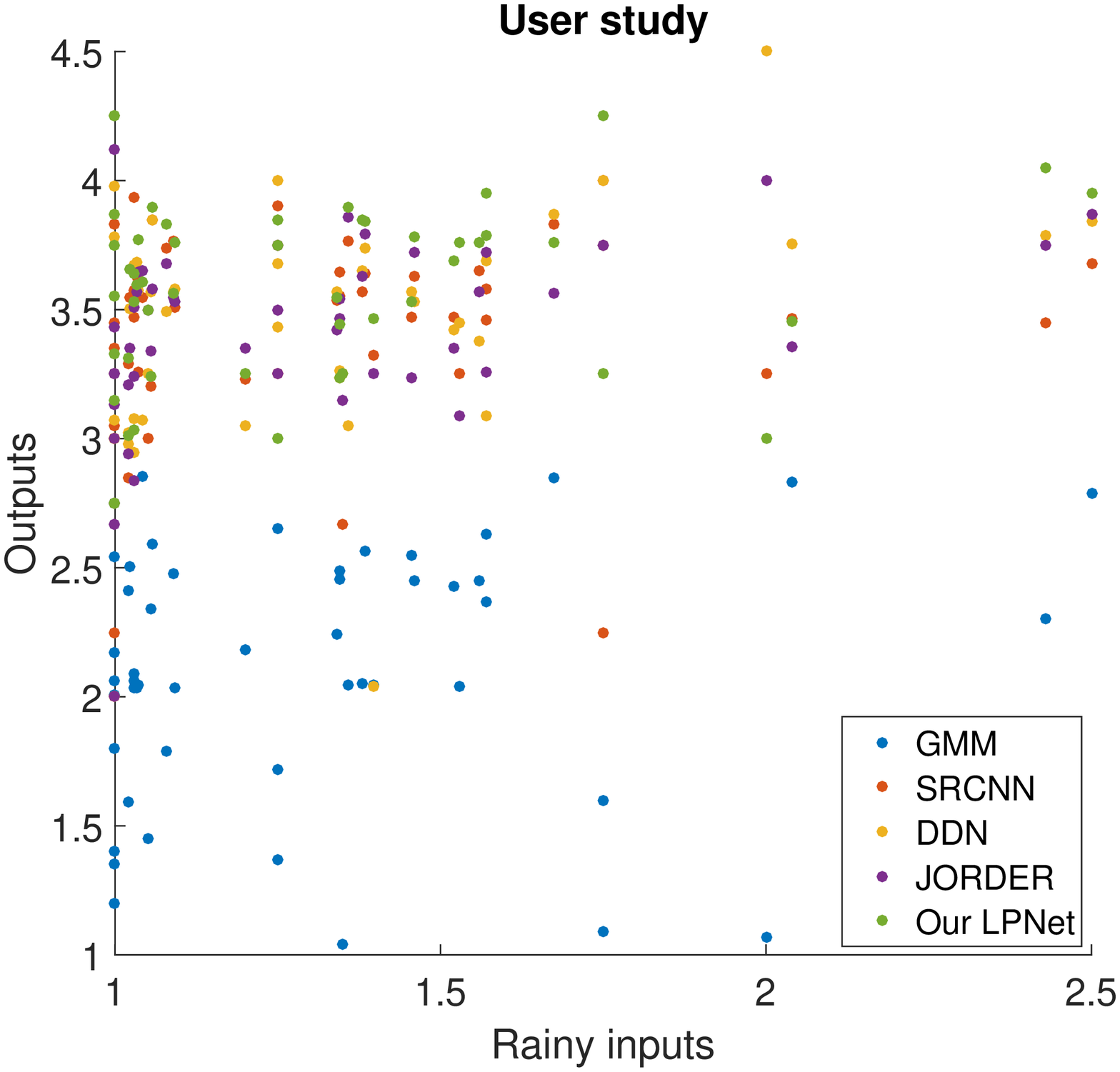}
\caption{Scatter plot of user study.} \label{fig.user}
\end{figure}

Since no ground truth exists, we construct an independent user study to provide realistic feedback and quantify the subjective evaluation. We collect 50 real-world rainy images from the Internet as a new dataset \footnote{Our code and data will be released soon.}. We use the compared five methods to generate de-rained results and randomly order the outputs, as well as the original rainy image, and display them on a screen. We then separately asked 20 participants to rank each image from 1 to 5 subjectively according to quality, with the instructions being that visible rain streaks should decrease the quality and clarity should increase quality (1 represents the worst quality and 5 represents the best quality). We show the average scores in Table \ref{tab.User} from these 1,000 trials and our LPNet has the best performance. In Figure \ref{fig.user}, we show the scatter plot of the rainy inputs vs de-rained user scores. This small-scale experiment gives additional support that our LPnet improves the de-raining on real-world images.
\begin{table}
\caption{Average scores of user study.}
\label{tab.User}
\centering
\begin{tabular}{|c|c|c|c|c|c|c|}
\hline
Inputs & GMM & SRCNN & DDN &JORDER&Our LPNet\\
\hline
 1.31 & 2.12  &3.39& 3.41 &3.41&3.58\\
\hline
\end{tabular}
\end{table}

\begin{figure}
\centering
\subfigure[Light rainy model]{\includegraphics[width = .23\textwidth]{3-our.jpg}}
\subfigure[Heavy rainy model]{\includegraphics[width = .23\textwidth]{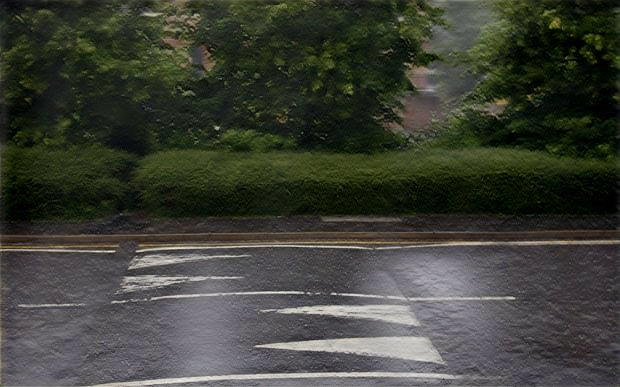}}
\caption{An example of dehazing effect. Our LPNet trained on the heavy rainy dataset can further improve image contrast.} \label{fig.dehaze}
\end{figure}

Moreover, when dealing with dense rain, LPNet trained on images with heavy rain has a dehazing effect as shown in Figure \ref{fig.dehaze}, which can further improve the visual quality. This is because the highest level sub-network (low-pass component) can adjust image contrast. Although dehazing is not the main focus of this paper, we believe that LPNet can be easily modified for joint deraining and dehazing.

\subsection{Running time and convergence}
To demonstrate the efficiency of LPNet, we show the average running time for a test image in Table \ref{tab:time}. Three different image sizes are chosen and each one is tested over 100 images. The GMM is implemented on CPUs according to the provided code, while other deep CNN-based methods are tested on both CPU and GPU. All experiments are performed on a server with Intel(R) Xeon(R) CPU E5-2683, 64GB RAM and NVIDIA GTX 1080. The GMM has the slowest running time since complicated inference is required to process each new image. Our method has a comparable and even faster computational time on both CPU and GPU compared with other deep models. This is because LPNet uses relatively shallow networks for each level, so requires fewer convolutions.
\begin{table*}
\centering
\caption{Comparison of running time (seconds).}
\begin{tabular}{|*{11}{c|}}
\hline
& \multicolumn{2}{|c|}{GMM \cite{Li2016Rain}} & \multicolumn{2}{|c|}{SRCNN \cite{dong2016image}}
& \multicolumn{2}{|c|}{DDN \cite{fu2017removing}} & \multicolumn{2}{|c|}{JORDER \cite{Yang2017Deep}}&  \multicolumn{2}{|c|}{Our LPNet}  \\\cline{1-11}
Image size         &CPU&GPU& CPU  & GPU & CPU & GPU & CPU  & GPU  & CPU  &GPU   \\\hline
500 $\times$ 500   &1.99$\times$10$^3$&-&0.25 &0.03 &1.51 &0.16 &2.95$\times$10$^2$  & 0.18 &0.67  &0.12 \\\hline
750 $\times$ 750   &3.09$\times$10$^3$&-&0.58 &0.09 &3.33 &0.22 &5.98$\times$10$^2$  & 0.36 &1.49  &0.16  \\\hline
1024 $\times$ 1024 &6.52$\times$10$^3$&-&1.07 &0.11 &5.40 &0.32 &1.20$\times$10$^3$  & 0.82 &2.46  &0.20  \\\hline
\end{tabular}
\label{tab:time}
\end{table*}

\begin{figure}
\centering
\includegraphics[width = 1\columnwidth]{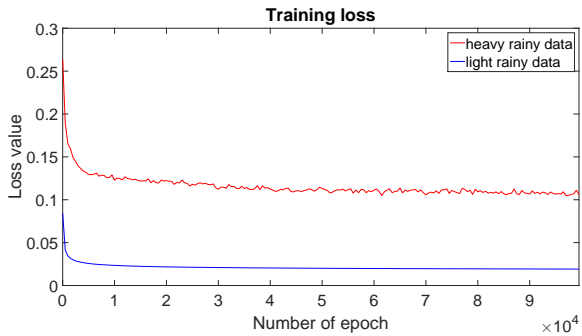}
\caption{Convergence on different training datasets.} \label{fig.loss}
\end{figure}

We also show the average training loss as a function of training epoch in Figure \ref{fig.loss}. We observe that LPNet converges quickly on training with both light and heavy rainy datasets. Since heavy rain streaks are harder to handle, as shown in the 1st row of Figure \ref{fig.syntheticH}, the training error of heavy rain streaks has a vibration.

\subsection{Parameter settings}
In this section, we discuss different parameters setting to study their impact on performance.
\subsubsection{Increasing parameter number}
\begin{figure}
\centering
\subfigure[Rainy image]{
\includegraphics[width = .15\textwidth]{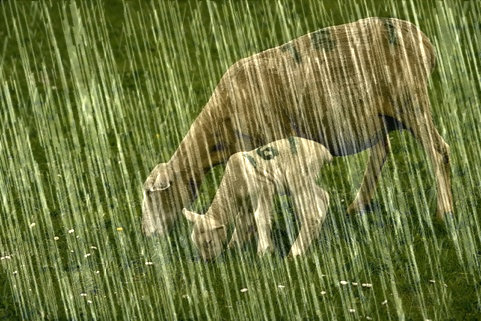}}
\subfigure[Default numbers]{
\includegraphics[width = .15\textwidth]{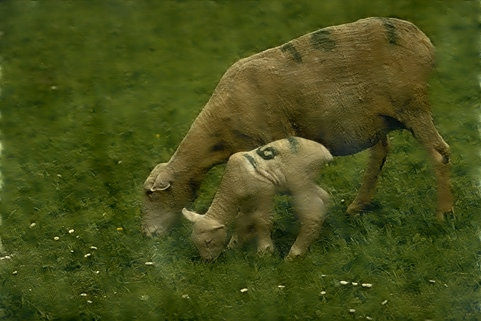}}
\subfigure[16 feature maps]{
\includegraphics[width = .15\textwidth]{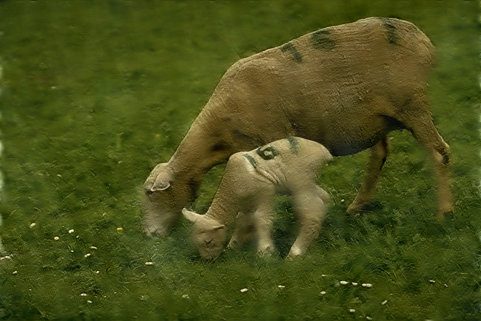}}
\caption{One example by using different parameter numbers.} \label{fig.numbers}
\end{figure}

We have conducted an experiment on the \emph{Rain100H} dataset with increased parameters, i.e., 16 feature maps for all convolution layers at each sub-network. The results are shown in Table \ref{tab.differentnumber}. As can be seen, the SSIM evaluation is better than JORDER and PSNR value is also improved. We believe that the performance can be further improved by using more parameters. However, increasing parameter number requires more storage and computing resources. Figure \ref{fig.numbers} shows one example by using different parameter numbers. As can be seen, the visual quality is almost the same. Thus, we use our diminishing parameter setting to achieve the balance between effectiveness and efficiency.
\begin{table*}
\centering
\caption{SSIM and PSNR value comparison for different parameters.}
\begin{tabular}{|*{7}{c|}}
\hline
& \multicolumn{2}{|c|}{JORDER \cite{Yang2017Deep}} & \multicolumn{2}{|c|}{Our LPNet (default)}
& \multicolumn{2}{|c|}{Our LPNet (increasing)} \\\cline{1-7}
 & SSIM &  PSNR  & SSIM    & PSNR & SSIM    & PSNR  \\\hline
\emph{Rain100H}   & 0.83 &\textbf{26.54} &0.81&23.73 &\textbf{0.84} &24.09\\\hline
Parameters \# &\multicolumn{2}{|c|}{369,792} & \multicolumn{2}{|c|}{7,548}
& \multicolumn{2}{|c|}{27,055} \\\cline{1-7}
\end{tabular}
\label{tab.differentnumber}
\end{table*}

\subsubsection{Skip connections}
\begin{figure}
\centering
\includegraphics[width = 1\columnwidth]{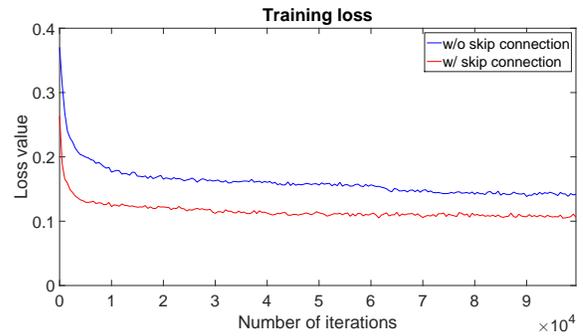}
\caption{Training curves w/ and w/o skip connections.} \label{fig.loss3}
\end{figure}

Though Laplacian pyramid images introduce sparsity in each level to simply the mapping problem, it is still essential to add skip connection in each sub-network. We adopt skip connection for two reasons. First, image information may be lost during feed-forward convolutional operations, using skip connection helps to propagate information flow and improve the deraining performance. Second, using skip connection helps to back-propagate gradient, which can accelerate the training procedure, when updating parameters. In Figure \ref{fig.loss3} we show the training curves on the heavy rainy dataset with and without all skip connections. As can be seen, using skip connection can bring a faster convergence rate and lower training loss.

\subsubsection{Loss function}
\begin{figure*}
\centering
\subfigure[Rainy image]{
\includegraphics[width = .3\textwidth]{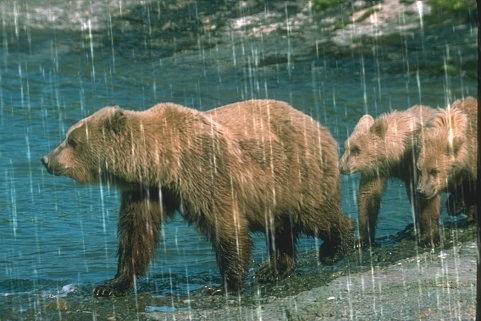}}
\subfigure[$\ell_2$ loss]{
\includegraphics[width = .3\textwidth]{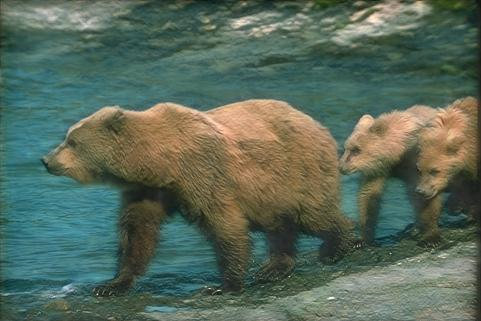}}
\subfigure[SSIM + $\ell_1$ loss]{
\includegraphics[width = .3\textwidth]{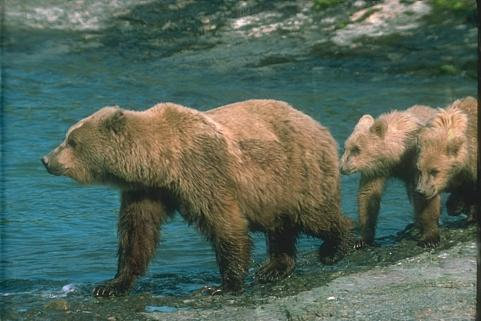}}
\caption{An deraining example by using different losses. Using SSIM + $\ell_1$ loss generates a more sharpen result.} \label{fig.loss1}
\end{figure*}

We use SSIM as a part of loss function (\ref{eq.overall}) for two main reasons. First, SSIM is calculated based on local image characteristics, e.g., local contrast, luminance and details, which are also the characteristics of rain streaks. Thus, using SSIM as the loss function is appropriate to guide the network training. Second, the human visual system is also sensitive to local image characteristics. SSIM has been motivated as generating more visually pleasing results, unlike PSNR. It has therefore become a more prominent measure in the image processing community. We also use $\ell_1$ loss because $\ell_1$ does not over-penalize larger errors and thus can preserve structures and edges. On the contrary, the widely used $\ell_2$ loss (which PSNR is based on) often generates over-smoothed results because it penalizes larger errors and tolerates small errors. Therefore, $\ell_2$ struggles to preserve underlying structures in the image compared with $\ell_1$. Figure \ref{fig.loss1} shows two results generated by using our combined loss (\ref{eq.overall}) and $\ell_2$ loss, respectively. As can be seen, using our combined loss (\ref{eq.overall}) can preserve more details.

\subsection{Extensions}
\subsubsection{Generalization to other image processing tasks}
\begin{figure}[t]
\centering
\includegraphics[width = .235\textwidth]{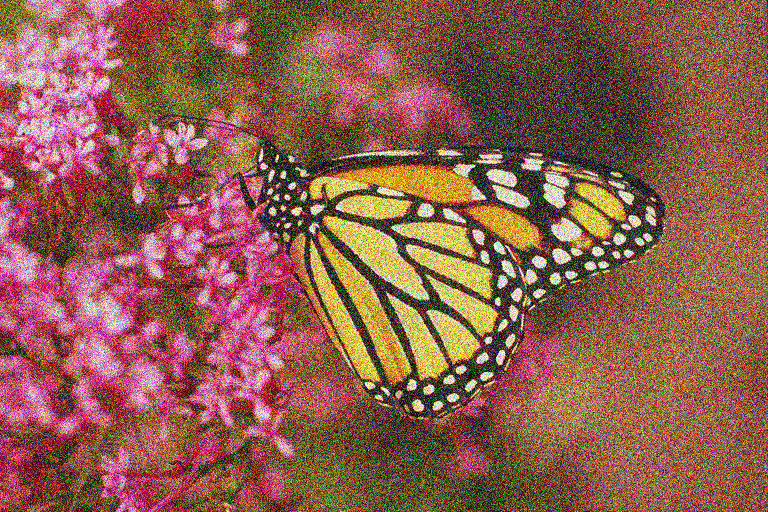}
\includegraphics[width = .235\textwidth]{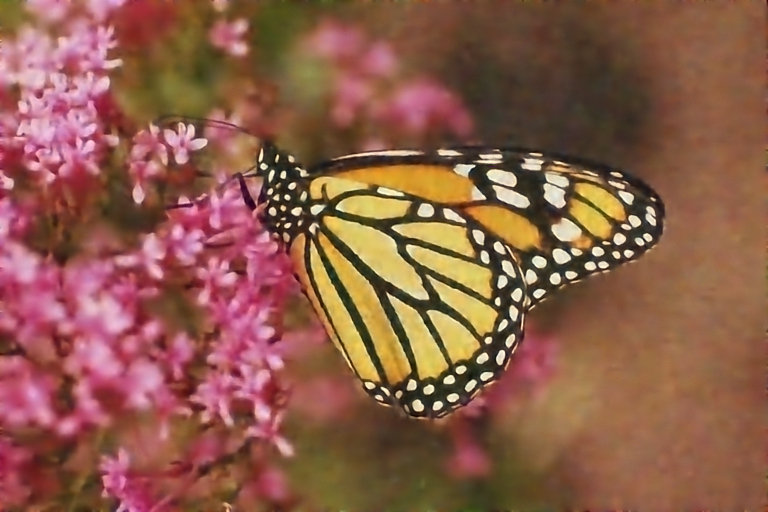}\\
\subfigure[Noise (top) and JPEG (bottom)]{\includegraphics[width = .235\textwidth]{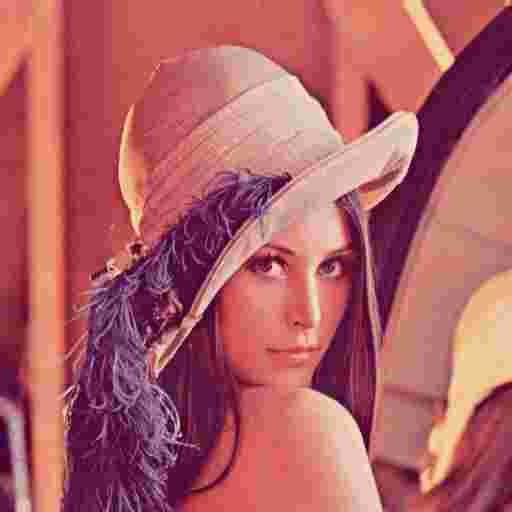}}
\subfigure[Our results]{\includegraphics[width = .235\textwidth]{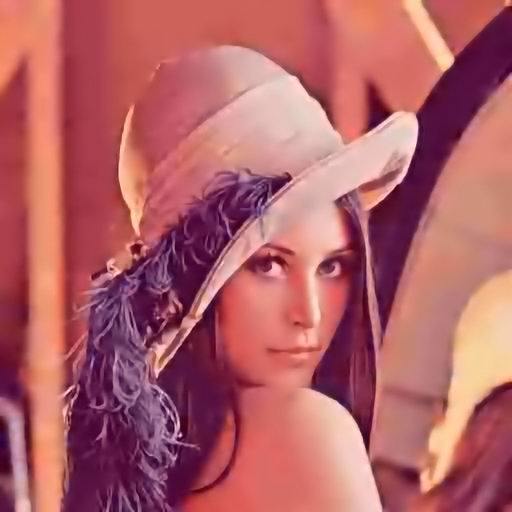}}
\caption{Denoising and reducing JPEG artifact.} \label{fig.exten}
\end{figure}

Since both Laplacian pyramids and CNNs are fundamental and general image processing technologies, our network design has potential value for other low-level vision tasks. Figure \ref{fig.exten} shows the experimental result on image denoising and JPEG artifacts reduction, which shares the property of rainy images in that the desired image is corrupted by high frequency content. This test demonstrates that LPNet can generalize to similar image restoration problems.

\begin{figure}
\centering
\includegraphics[width = 1.7in]{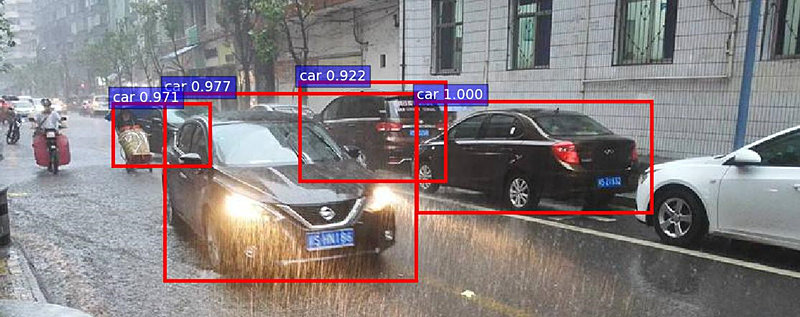}
\includegraphics[width = 1.7in]{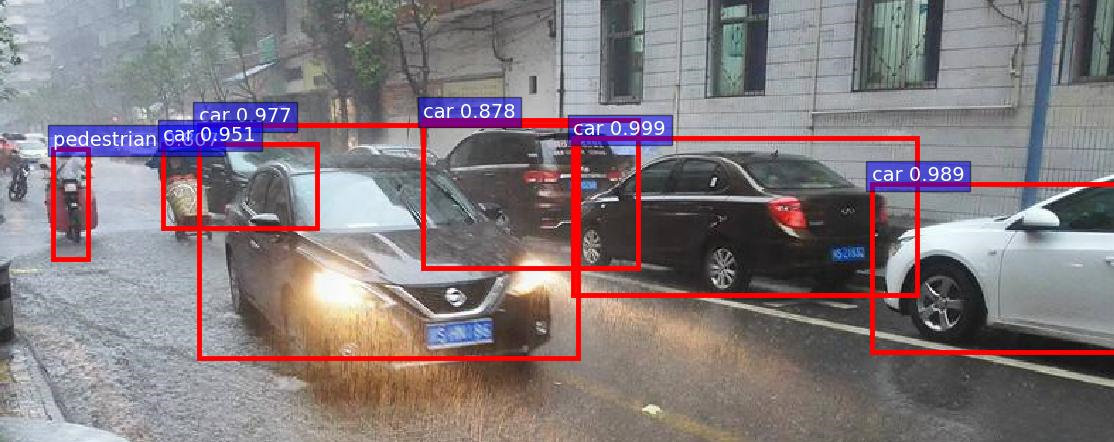} \\
\subfigure[Direct detection]{\includegraphics[width = 1.7in]{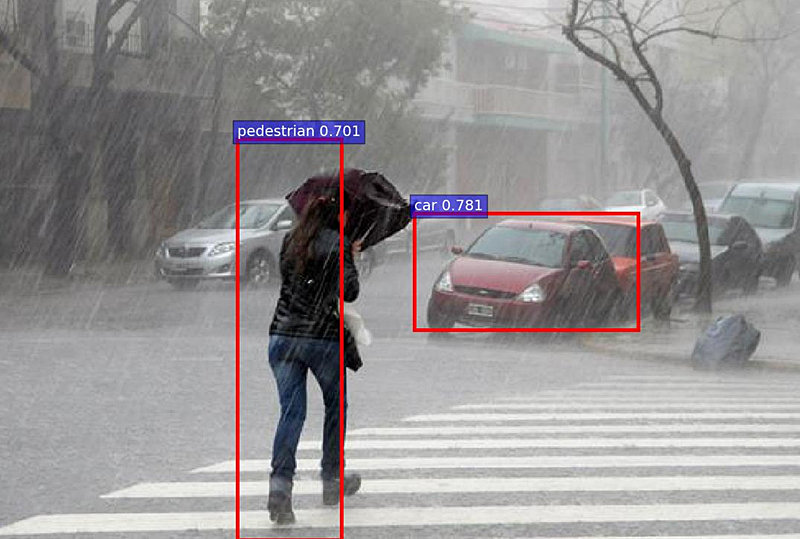}}
\subfigure[Deraining + detection]{\includegraphics[width = 1.7in]{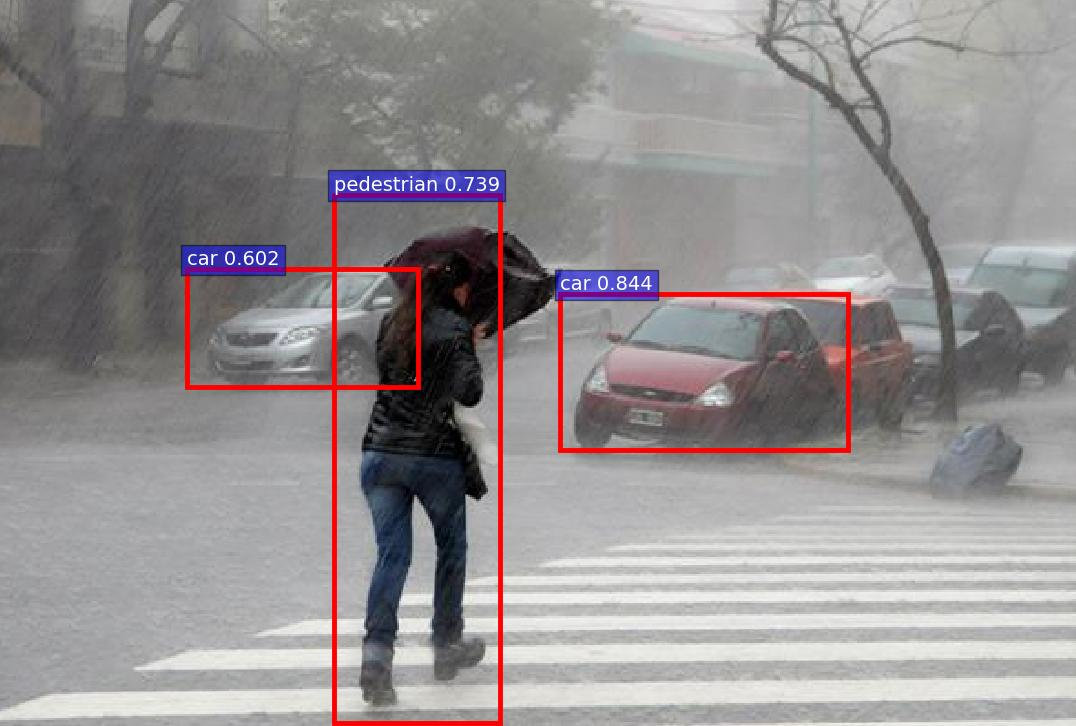}}
\caption{An example of joint deraining and object detection on a real-world image. We use the Faster R-CNN \cite{Ren2015Faster} to perform object detection with a confidence threshold of 0.8.} \label{fig.detect}
\end{figure}
\subsubsection{Pre-processing for high-level vision tasks}
Due to the lightweight architecture, our LPNet can potentially be efficiently incorporated into other high-level vision systems. For example, we study the problem of object detection in rainy environments. Since rain steaks can blur and block objects, the performance of object detection will degrade in rainy weather. Figure \ref{fig.detect} shows a visual result of object detection by combining with the popular Faster R-CNN model \cite{Ren2015Faster}. It is obviously that rain streaks can degrade the performance of Faster R-CNN, i.e., by missing detections and producing low recognition confidence. On the other hand, after deraining by LPNet, the detection performance has a notable improvement over the naive Faster-RCNN.

Additionally, due to the lightweight architecture, using LPNet with Faster R-CNN does not significantly increase the complexity. To process a color image with size of $1024\times1024$, the running time is 3.7 seconds for Faster R-CNN, and 4.0 seconds for LPNet + Faster R-CNN.

\section{Conclusion}
In this paper, we have introduced a lightweight deep network that is based on the classical Gaussian-Laplacian pyramid for single image deraining. Our LPNet contains several sub-networks and inputs the Laplacian pyramid to predict the clean Gaussian pyramid. By using the pyramid to simplify the learning problem and adopting recursive blocks to share parameters, LPNet has fewer than $8$K parameters while still achieving good performance. Moreover, due to the generality and lightweight architecture, our LPNet has potential values for other low- and high-level vision tasks.

\bibliographystyle{IEEEbib}
\bibliography{LPNet}

\end{document}